\documentclass{article}

\PassOptionsToPackage{numbers}{natbib}

\usepackage{enumitem}

\usepackage{makecell}

    \usepackage[final]{neurips_2021}

\usepackage[utf8]{inputenc} 
\usepackage[T1]{fontenc}    
\usepackage{hyperref}       
\usepackage{url}            
\usepackage{booktabs}       
\usepackage{amsfonts}       
\usepackage{nicefrac}       
\usepackage{microtype}      
\usepackage{xcolor}         
\usepackage{graphicx}

\usepackage{authblk}

\newcommand{\Fc}{{F3}}
\newcommand{\Fd}{{F4}}

\title{Passive Attention in Artificial Neural Networks \\ Predicts Human Visual Selectivity}

\author[1,b,c]{\textbf{Thomas~A.~Langlois}}
\author[1,b]{\textbf{H.~Charles~Zhao}}
\author[d]{\textbf{Erin~Grant}}
\author[e]{\textbf{Ishita~Dasgupta}}
\author[2,a,b]{\textbf{Thomas~L.~Griffiths}}
\author[2,c]{\textbf{Nori~Jacoby}}

\affil[1]{T.A.L. and H.C.Z. contributed equally to this work.}
\affil[2]{T.L.G. and N.J. contributed equally to this work.}

\affil[a]{Department of Psychology, UC Berkeley}
\affil[b]{Department of Computer Science, Princeton University}
\affil[c]{Computational Auditory Perception Research Group, Max Planck Institute for Empirical Aesthetics}
\affil[d]{Department of Electrical Engineering and Computer Sciences, UC Berkeley}
\affil[e]{DeepMind, New York}

\begin{document}

\maketitle

\begin{abstract}

Developments in machine learning interpretability techniques over the past decade have provided new tools to observe the image regions that are most informative for classification and localization in artificial neural networks (ANNs). Are the same regions similarly informative to human observers? Using data from 79 new experiments and 7,810 participants, we show that passive attention techniques reveal a significant overlap with human visual selectivity estimates derived from 6 distinct behavioral tasks including visual discrimination, spatial localization, recognizability, free-viewing, cued-object search, and saliency search fixations. We find that input visualizations derived from relatively simple ANN architectures probed using guided backpropagation methods are the best predictors of a shared component in the joint variability of the human measures. We validate these correlational results with causal manipulations using recognition experiments. We show that images masked with ANN attention maps were easier for humans to classify than control masks in a speeded recognition experiment. Similarly, we find that recognition performance in the same ANN models was likewise influenced by masking input images using human visual selectivity maps. This work contributes a new approach to evaluating the biological and psychological validity of leading ANNs as models of human vision: by examining their similarities and differences in terms of their visual selectivity to the information contained in images.
\end{abstract}

\section{Introduction}

The last decade has witnessed the rise of artificial neural networks (ANNs) that can match and even exceed human performance on a variety of perceptual and cognitive tasks, ranging from image recognition \cite{krizhevsky2012imagenet} to natural language processing and reinforcement learning \cite{lecun2015deep}. Alongside the rapid development of these technologies, a significant body of work aimed at improving the interpretability of these systems and comparing them to biological ones has also grown \cite{yamins2014performance, kell2018task, yamins2020optimization, linsley2018learning,linsley2017visual}. In computer vision, techniques for probing which visual regions ANNs ``attend to'' when classifying images have been developed to visualize the receptive fields of convolutional layers as well as regions of a visual input that most influence the class activations of the models \cite{xu2015show,zeiler2014visualizing,zhou2016learning,mnih2014recurrent,olah2017feature}. In neuroscience, researchers began to quantify the functional fidelity of leading ANNs as models of the human visual system using both neural and behavioral benchmarks \cite{kubilius2019brain}. Finally, cognitive scientists have developed techniques to compare the structure of ANN learned representations to human psychological representations \cite{battleday2020capturing, peterson2019learning}. All these efforts have contributed to our understanding of the biological and psychological validity of leading ANNs as models of biological vision, beyond just assessing their performance on standard object categorization benchmarks. 

Many previous analyses of the correspondence between ANNs and human vision have focused on the {\em  representations} used by the systems.
However, a natural question is whether ANNs {\em select} information in the same way, and in particular whether they attend to the same visual regions as humans  when extracting information for visual object recognition and localization. Prior work has developed ANNs trained explicitly to predict human visual gaze \cite{akinyelu2020convolutional}, and even incorporated simulated foveated systems into the model design \cite{deza2020emergent}. In addition, work comparing human attention to computational attention \cite{lai2020understanding, das2017human, ebrahimpour2019humans, linsley2018learning,linsley2017visual} revealed that computational attention tends not to resemble human attention. However, this work also provides evidence that using human attention to supervise the training of network architectures devised to emulate the parallel pathways in the human visual system can improve performance, and yields more human-like visual features \cite{linsley2018learning}. Still, relatively little work has attempted a more comprehensive examination of how a wide range of ANNs compare to multiple human behavioral measures using a larger variety of interpretability techniques that are now available to probe what visual information ANNs use.

Methods for gaining insight into what is learned by ANNs started with ``passive'' attention gradient-based approaches designed to reveal which visual inputs influence the class activation score the most \cite{simonyan2013deep}. More advanced techniques using deconvolution and guided backpropagation methods followed \cite{zeiler2014visualizing,springenberg2014striving} as well as techniques that introduced novel design alterations, such as global average pooling layers and class activation mapping (\textsc{cam}) to localize class-specific visual regions in the input images \cite{zhou2016learning}. Finally, more general approaches that could be applied to architectures without global average pooling \cite{selvaraju2017grad} appeared, with some of the most recent contributions proposing class activation mapping techniques that do not rely on gradients at all \cite{wang2020score}. Aside from this range of ``passive'' techniques, computer scientists have also developed CNN models that incorporate end-to-end trainable attention modules \cite{jetley2018learn,fukui2019attention} both as a means for improving interpretability and boosting performance. The full range of techniques now available for visualizing the information that is most relevant to ANNs offer an unprecedented and unique opportunity to compare their results to biological analogues such as estimates of human attention, discrimination accuracy, and visual recognition over image regions.     

Since the early 19th century \cite{finger2001origins} vision scientists devoted to the study of biological vision have also developed a variety of experimental techniques for estimating the visual information used by the primate visual system when engaged in similar perceptual and cognitive tasks such as visual search, localization, and recognition. Among these are measures of visual change sensitivity (discrimination accuracy), visuospatial memory (spatial localization estimation), as well as explicit reports of visual recognizability. In this work, we obtained estimates from six different perceptual tasks, including explicit visual recognizability estimates using a recent behavioral task \cite{henderson2018meaning,henderson2017meaning} (Fig. \ref{fig:Tasks}A and Fig. \ref{fig:Human-Maps}A), a two-alternative forced choice (2AFC) change sensitivity task (Fig. \ref{fig:Tasks}B and Fig. \ref{fig:Human-Maps}A), eye-tracking fixations (Fig. \ref{fig:Tasks}C and Fig. \ref{fig:Human-Maps}A), and a spatial localization task (Fig. \ref{fig:Tasks}D and Fig. \ref{fig:Human-Maps}A).

We then compared them to estimates of ANN visual selectivity using a variety of pretrained models and visualization techniques including guided backpropagation and techniques based on class activation mapping. We used a range of model types, including early convolutional networks like AlexNet \cite{krizhevsky2012imagenet}, as well as recent state-of the-art models, such as Visual Transformer (ViT) \cite{dosovitskiy2020image} and EfficientNet \cite{tan2019efficientnet} models. 

We find that only a select class of ANN models and passive attention techniques capture the shared variance across all human visual selectivity measures. This work contributes to current efforts aimed at evaluating the biological and psychological validity of contemporary ANNs by investigating the similarity between artificial and biological vision systems at the level of the visual inputs rather than the learned representations or correspondence to patterns of neural activation in the visual cortex \cite{kubilius2019brain}. We see our primary contributions as two-fold:
(1) In a departure from prior work, we use a larger range of human behavioral measures, ANN models, and attention techniques in our comparisons. These reveal a significant range in the degree to which different models combined with different attention techniques produce human-like results. In addition, they show that more human-like ANN attention captures a shared component of the joint variability between the different human measures, rather than any particular human measure. (2) We provide causal evidence that more human-like ANN attention directly influences both human classification performance as well as predictions made by ANNs.

\section{Human visual selectivity measures}
\label{sec:human-measures}

\begin{figure}[!htb] 
    \centering
        \includegraphics[width=\linewidth]{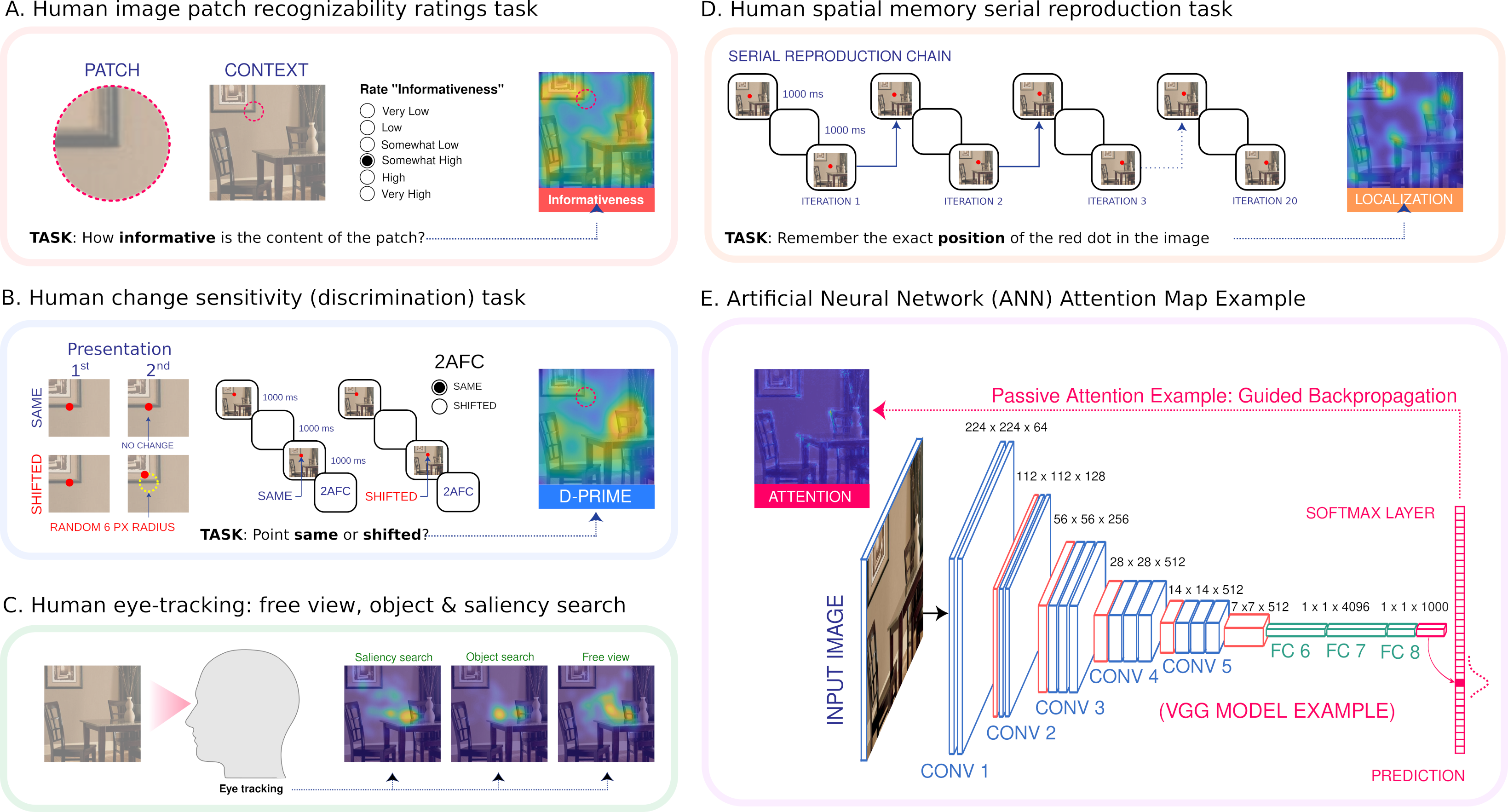}
    \caption{Human behavioral tasks, and ANN attention. A. Informativeness patch ratings task. B. Discrimination accuracy 2AFC task. C. Eye-tracking for free search, object search, and saliency search. D. Spatial memory serial reproduction task. E. ANN attention (passive attention example). Details of the experimental procedures and map estimation are included in Appendix Fig. 1.}
    \label{fig:Tasks}
\end{figure}

We computed 6 behavioral measures with 25 images, performing experiments in which a total of 4,050 participants took part (see Table 1 in the Appendix). We ran experiments for 3 distinct human behavioral tasks for each image (see Fig. \ref{fig:Tasks} and Fig. \ref{fig:Human-Maps} for representative examples). A total of 1,575 participants took part in the discrimination accuracy experiments (an average of 63 participants for each of the 25 images), and a total of 225 participants took part in each of the image patch ratings tasks (9 participants per image). Finally, the spatial memory serial reproduction chain experiments were completed by a total of 2,250 participants (an average of 90 participants for each of the images). Participants were recruited anonymously over Amazon Mechanical Turk (AMT), and provided informed consent. Participants were paid approximately \$7 per hour. Details of the experimental procedures, design, and map estimation for each task can be found in Appendix Fig. S1.

\begin{enumerate}[wide, labelindent=0pt]
  \item \textbf{Visual recognizability.} We adopted a recent behavioral task \cite{henderson2018meaning,henderson2017meaning} designed to measure the informativeness and recognizability of local image regions by using explicit self-reports (see Fig. \ref{fig:Tasks}A). In the task, participants view small circular image patches sampled from full images and rate how ``recognizable'' or ``informative'' the content of the patch is on a six-point Likert scale ranging from a rating of ``Very low recognizability'' to ``Very high recognizability''. The patches were sampled from a regular 12 x 12 grid over the image. Fig. \ref{fig:Human-Maps}A shows representative examples of the results following averaging over all the ratings in different spatial areas of the images, smoothing, and interpolation to produce continuous maps for each image. We ran 25 experiments with an average of 9 participants per image (see Appendix for details of the map generation procedure). A total of 225 participants took part in the patch ratings experiments. Participants were paid \$2 to complete 144 experimental trials. 
  \item \textbf{Change sensitivity (Discrimination).} We measured change sensitivity using a two-alternative forced choice (2AFC) discrimination task. In this task, participants viewed an image with a small red dot superimposed on it for 1000 milliseconds. Following a 1000 millisecond delay, the same image was presented again with the dot in either the same exact location or in a slightly displaced location (see Fig \ref{fig:Tasks}B). Participants were then asked to indicate if the dot was shifted or unchanged in the second presentation. Crucially, the locations of the initial dot locations were sampled densely from all possible locations on a regular grid that spanned the dimensions of the image, in order to measure changes in visual change sensitivity over the entire image (see Appendix for details). This task measures changes in visual acuity conditioned on different visual areas in an image, and has been used as a proxy for measuring variable encoding precision of different image regions \cite{langlois2021serial}.  We ran 25 experiments (one for each image) with an average of 63 participants per experiment. The overall number of participants for the discrimination tasks equalled 1,575. Participants were paid \$1.50 to complete 120 trials in the discrimination task.
  \item \textbf{Visuospatial localization.} We used a recent behavioral paradigm based on serial reproduction that can reveal intricate spatial memory priors that guide visual localization estimation in humans \cite{langlois2021serial}. In this paradigm the first participant views a point superimposed on an image and then reproduces its location from memory. The next participant views the same image but with the point located in the position reconstructed by the previous participant. As in the ``telephone game,'' the process is repeated, forming a chain of participants. For each image, there were a total of 20 iterations in the chains, for 250 initial random seed dot positions. This experimental procedure is known to reveal the spatial landmarks in visual scenes that bias human allocentric visuospatial representations (see \cite{langlois2021serial} and Fig. \ref{fig:Tasks}D). We ran 25 experiments (one for each image) with an average of 90 participants each. The overall number of participants was 2,250. Participants were paid a base rate of \$1.00 for completing 105 trials in the spatial memory experiment but could earn up to \$1.50 depending on accuracy in the task. Additional details are provided in the Appendix.
  \item \textbf{Fixations.} Finally, we used an existing dataset of human fixations obtained via eye-tracking when human participants were engaged in a free-viewing task, a cued object search task, and a saliency search task \cite{koehler2014saliency} (Fig. \ref{fig:Tasks}C). We used the published data from the 75 experiments reported in \cite{koehler2014saliency}. 
\end{enumerate}

\section{ANN models, passive and active attention}
We evaluated three standard deep convolutional neural network architectures (AlexNet \cite{krizhevsky2012imagenet}, VGGNet \cite{simonyan2013deep}, ResNet \cite{he2016deep}), as well as two state-of-the-art architectures (Vision Transformer (ViT) \cite{dosovitskiy2020image} and EfficientNet \cite{tan2019efficientnet}). For each of these models, which do not have active attention modules, we obtained attention maps using a range of passive methods described below. We also evaluated two built-in end-to-end trainable attention modules, described below. All models were pretrained on ImageNet 2012 \cite{deng2009imagenet}, CIFAR-100 \cite{krizhevsky2009learning}, or Places365-Standard \cite{zhou2017places}. Unlike the ImageNet and CIFAR-100 datasets, which are comprised of images of objects, Places365-Standard is comprised of images of complex natural scenes. See Table 2 in the Appendix for a list of all models.
\subsection{Passive attention}
We used gradient-based techniques including guided backpropagation methods, as well as more recent techniques based on class activation mapping (see Fig. \ref{fig:Tasks}E for a schematic example of passive attention). The methods based on guided backpropagation effectively try to compute the sensitivity of the model's output with respect to each pixel in the input image, using various techniques for increasing the signal in these maps and decreasing noise. These methods include guided backpropagation (GBP) \cite{springenberg2014striving}, guided gradients times the image (GBPxIM) \cite{shrikumar2016not}, and SmoothGrad with guided backpropagation (SGBP) \cite{smilkov2017smoothgrad}. The methods based on class activation mapping compute linear combinations of the activation maps in the final convolutional layer of the model in order to determine the discriminative regions of the image used by the model. These methods include Grad-CAM \cite{selvaraju2017grad}, Score-CAM \cite{wang2020score}, and CAMERAS \cite{jalwana2021cameras}. See the Appendix for details on these passive attention methods.
\subsection{Active attention}
We evaluated two different active attention modules. These active attention modules are trainable, and they learn to generate masks which are applied to the input image (or to intermediate convolutional layers). These masks are effectively explicit attention maps, so we do not have to use passive attention methods to try to discern the models' attention. These attention modules can be incorporated into essentially any standard CNN architecture, so we chose a few for which we were able to obtain pretrained weights (see the Appendix for details).

One attention module is described in the paper Learn to Pay Attention \cite{jetley2018learn}, which we will refer to as LTPA. LTPA inserts attention at three intermediate convolutional layers within a VGGNet architecture. The other active attention module is used in Attention Branch Networks (ABNs) \cite{fukui2019attention}. ABN has a separate ``attention branch'' that runs in parallel with a ``perception branch,'' and is based on class activation mapping \cite{zhou2016learning}.

\section{Passive attention predicts shared variance across human measures}

\begin{figure}[!htb] 
    \centering
        \includegraphics[width=\linewidth]{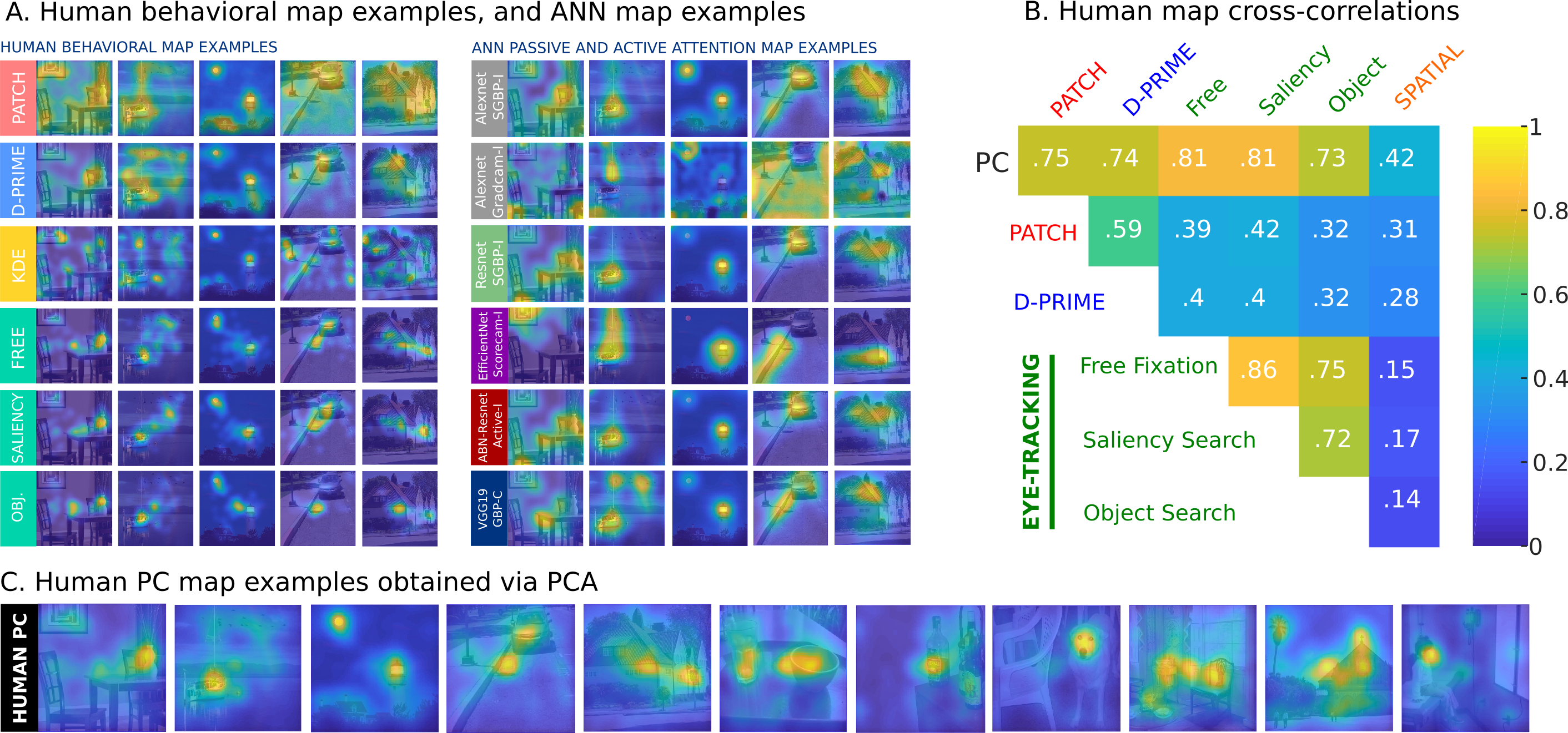}
    \caption{Human behavioral task maps and ANN maps. A.  Representative examples of the human maps obtained for the (1) patch ``informativeness'' ratings task (2) The discrimination accuracy task (3) the serial reproduction spatial memory task (4) Free fixations (5) saliency search fixations and (6) cued object search fixations. Examples of ANN maps for the same images are also shown. B. Cross correlations of all the maps, including the linear combination of all the maps that predicts the maximal shared variance (Human PC), are shown. While most methods are relatively highly intercorrelated, there are clear differences. Fixations were the most highly intercorrelated (\textit{r} = .72-.86), while spatial memory Kernel Density Estimates (KDEs) are only weakly correlated (\textit{r} = .14-.17) to the fixations, in line with previous findings \cite{langlois2021serial}. Factor loadings of each of the six measures to the Human PC were uniformly high (\textit{r} = .42-.81). C. Human PC map examples.}
    \label{fig:Human-Maps}
\end{figure}

\textbf{Human experiments.} Fig. \ref{fig:Human-Maps}A shows representative results from each of the human behavioral experiments. We found that the human maps were correlated with one another though  the correlations varied from \textit{r} = 0.14 to 0.86 (see Fig. \ref{fig:Human-Maps}B) due to variations in the visual regions that were the most implicated from one task to another.  We obtained a single factor that captures the maximal amount of shared variance across all behavioral maps (``Human PC''; see Fig. \ref{fig:Human-Maps}C) by computing a linear combination of the results of the six experiments via Principal Component Analysis (see SI Appendix for details and formal description). The Human PC was correlated with each of the six behavioral measures (\textit{r} = .75, .74 , .42, .81, .81 and .73 for the informativeness maps, change sensitivity maps, spatial localization maps, and each of the fixation maps, respectively).

\textbf{ANN maps.}  We then compared the maps computed by the ANN attention methods (Fig. \ref{fig:ANN-Human}A; Raw ANN maps are included in Appendix Fig. S3). We found that ANN maps varied significantly in terms of their level of ``smoothness'' depending on the attention method. Because of this, and in order to compare the human and ANN maps in a way that is agnostic to the raw smoothness of the ANN maps, we introduced a smoothing parameter when comparing the two. We optimized the smoothing parameter for each of the ANN maps based on the correlation of the result to each of the human behavioral maps including the Human PC. We did this by applying the same smoothing to each of the individual 25 image ANN maps, and then computing the average Pearson correlation of those maps to the corresponding human maps for each of the 25 images. We repeated this process for each ANN attention method, and for each human task.

\begin{figure}
    \centering
        \includegraphics[width=\linewidth]{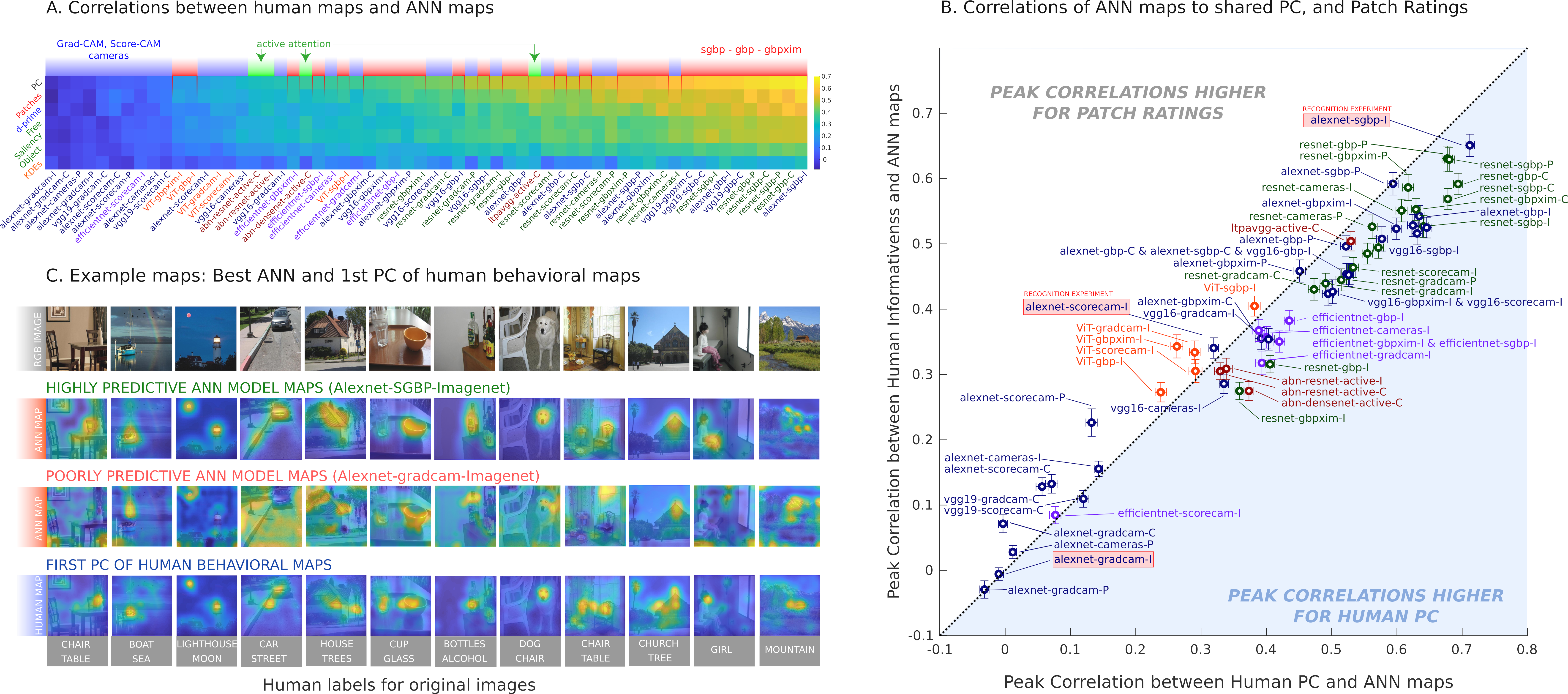} 
    \caption{Human behavioral maps and ANN attention. A. Cross-correlations between each of the human maps (including the shared Human PC) and the ANN maps. Results are sorted according to the average peak correlations across the six human measures and PC maps. Blocks highlighted in red correspond to ANN maps obtained using SGBP, GBP, or GBPxIM methods. Blocks highlighted in green correspond to maps obtained from LTPA and ABN active attention models. Blocks with blue highlights correspond to maps obtained using CAM-based methods. Results show that maps obtained using GBP-based attention methods are better aligned with human behavioral maps than both CAM-based results and active attention results (\textit{p} < 0.001, and \textit{p} < 0.001). The naming convention for passive attention maps is <architecture>-<passive attention method>-<I/C/P for ImageNet/CIFAR-100/Places365>. The naming convention for active attention maps is <attention module>-<architecture>-active-<I/C for ImageNet/CIFAR-100>. B. Correlations between all ANN maps and Human PC (x-axis) and the correlation between all ANN maps and the Human Informativeness ratings.  Error bars were estimated from 100 bootstrapped samples of the human data. The red boxes indicate the ANN maps used for the human speeded recognition task. The blue shaded area represents the scenario where the Human PC correlations to ANN maps (optimized to Human PC maps) are higher than the corresponding correlations between human informativeness maps and ANN maps (optimized to human informativeness maps). C. Representative examples of the ANN maps that were highly predictive of the Human PC maps (\textit{r} = 0.71, \textit{p} < 0.001), and not highly predictive of the Human PC maps (\textit{r} = 0) are shown. Also shown are the Human PC maps for the same images.}
    \label{fig:ANN-Human}
\end{figure} 

\textbf{The relation between ANN and human maps.} Next, we explored the relation between each of the human measures (including the Human PC) and the ANN attention maps. Fig. \ref{fig:ANN-Human}A shows the optimal correlations between human and ANN maps for each human map type including the Human PC. The peak correlations between the Human PC maps and the most highly correlated ANN maps exceed the peak correlations of the same ANN maps to each of the six behavioral maps (Fig. \ref{fig:ANN-Human}B; PC average \textit{r} = 0.71, compared with \textit{r} = 0.36-0.65 for each of the human maps). This suggests that the peak ANN maps predict the shared component of the variance between all the human measures rather than the variance of any particular human map type (such as human fixations or discrimination accuracy for instance). This is significant because it indicates that some intrinsic aspects of visual information contained in the images captured by the peak ANN models are predictive of human visual selectivity regardless of the behavioral task, and in spite of the systematic differences between them \cite{langlois2021serial}. Fig. \ref{fig:ANN-Human}B illustrates this fact for the human patch ratings task results. For most ANN maps, the correlation values under the diagonal in the plot indicate that the peak correlations of the ANN models to the Human PC were significantly higher than the peak correlations of the same ANN models to the Human patch ratings results. Fig. S4 illustrates the same finding for all individual human behavioral measures. In addition, Fig. \ref{fig:ANN-Human}B shows the top correlations for all ANN maps reaching a peak of \textit{r} = 0.71 for the SGBP method applied to the AlexNet network pretrained on ImageNet (alexnet-SGBP-I). In a separate analysis, we repeated the smoothing parameter fitting using split-half cross-validation, and found that performance of smoothed hold-out test set maps using smoothing parameters fit to random training set maps produced nearly identical ranges in peak correlations to the human PC (between \textit{r} = 0.73 and \textit{r} = -0.01 for the training set, and between \textit{r} = 0.72 and \textit{r} = -0.04 for the testing set), as well as a nearly identical rank order in peak correlations to the human PC. Details of this analysis are included in the SI Appendix (see Fig. S\ref{fig:S7}).

A natural question concerns whether the peak correlations achieved by the leading ANN maps are due to the model architecture, attention method, or training set. Fig. \ref{fig:ANN-Human}A shows correlations of the ANN maps to the human measures by attention type. We used the average correlations across the 6 measures and images as a dependent variable and found that the attention method category explains 35.9\% of the variance, while architecture and training set categories explain 25.3\% and 1.5\% of the variance, respectively (see Fig. \ref{fig:ANN-Human}A for the average performance across all map types, and for each of the human map types). This result suggests that attention type plays the most significant role in our findings.  Surprisingly, SGBP applied to one of the simplest architectures (the AlexNet network pretrained on ImageNet) showed the highest peak correlation to the Human PC (\textit{r} = 0.71), and was significantly more predictive of the Human PC and human patch ratings than most of the other ANN maps (\textit{p} < 0.001 with Bonferroni corrections applied). 

Passive attention models were consistently predictive of the human maps across behavioral tasks (See Appendix Fig. S2, S3 and S4). However, there was significant variation in the performance of these methods. For the human PC results, Guided-backpropagation based attention results (GBP, SGBP and GBPxIM; averaged \textit{r} = 0.51) consistently outperformed active attention results (averaged \textit{r} = 0.38, \textit{p} = 0.001), and these in turn outperformed Class Activation Mapping (CAM) based attention results (Grad-CAM, Score-CAM and CAMERAS, averaged \textit{r} = 0.27, \textit{p} = 0.001). In addition, we found that the resnet models (averaged \textit{r} = 0.54) produced more human-like maps than the remaining models (averaged \textit{r} ranged between 0.25 - 0.42, \textit{p} < 0.001 in all cases; with the Bonferroni correction applied). Finally, comparing results from the model types for which we could extract both active and passive attention maps reveals that guided-backpropagation based attention produces more human-like results than active attention architectures (See SI Appendix for details).

\section{Validation experiment: ANN visual selectivity boosts human recognition}
\label{sec:human-recog}

\begin{figure}[!htb] 
    \centering
        \includegraphics[width=\linewidth]{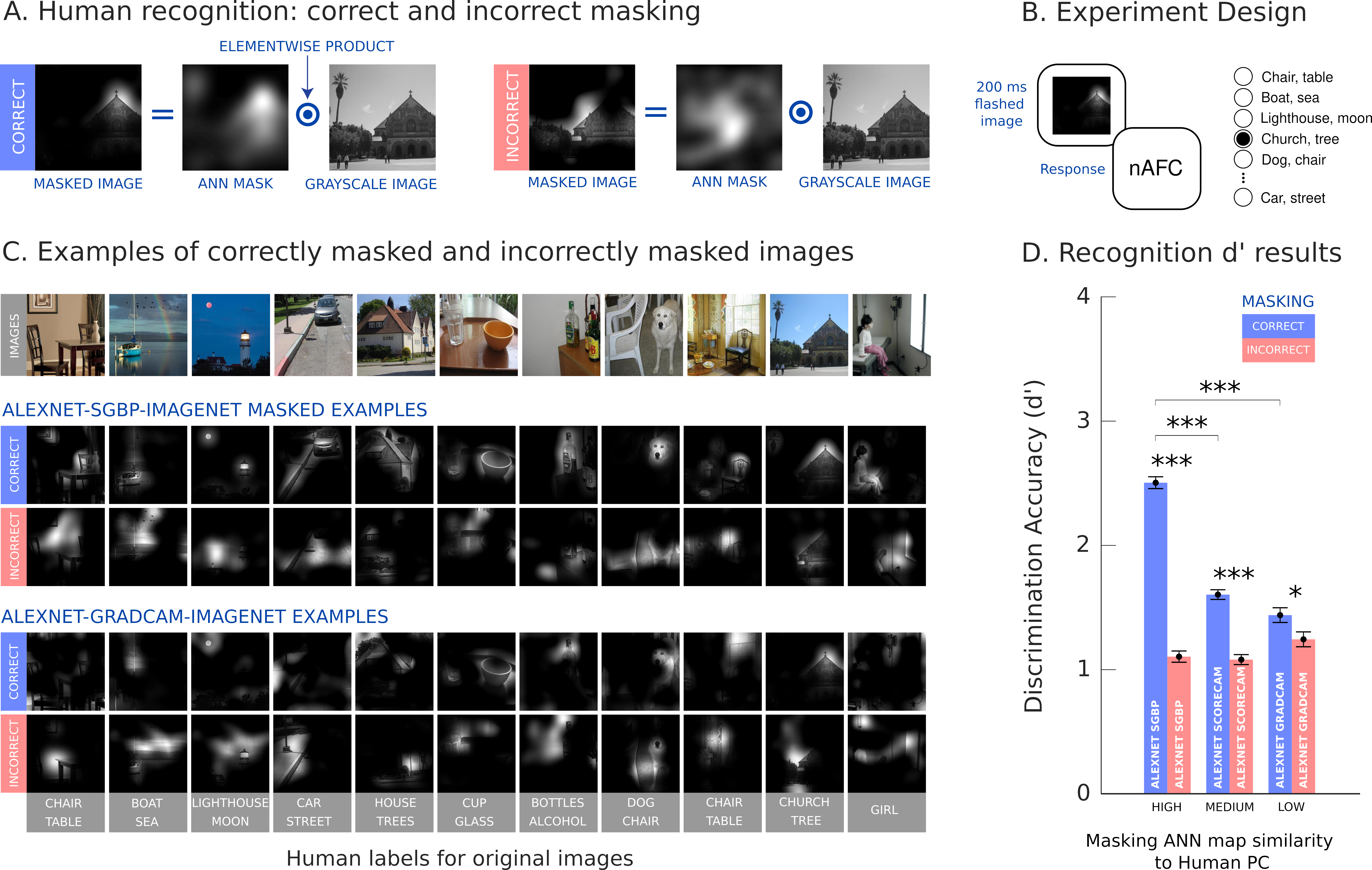}
    \caption{Human recognition results. A. Masking procedure. We masked images by an elementwise product of the grayscale image with the ANN map. B. Experimental design. Participants viewed a masked image for 200 milliseconds. They then selected the word descriptors that best described the image. C. Example original images, and corresponding masked versions obtained from the ImageNet-pretrained alexnet-SGBP maps (high peak correlation to Human PC), and from the same model also pretrained on ImageNet using Grad-CAM attention (low peak correlation to Human PC). D. Overall $d'$ results for the experiments using SGBP, Score-CAM, and Grad-CAM maps. Results reveal significantly higher $d'$ for correctly masked images (blue bar) relative to incorrectly masked images (red bar) for all 3 ANN map types (\textit{p} < 0.001, \textit{p} < 0.001, \textit{p} < 0.05 for SGBP, Score-CAM, and Grad-CAM maps, respectively). In the correct masking condition (blue bars), there was a significant difference between $d'$ results for the SGBP maps relative to the $d'$ results for the Score-CAM and Grad-CAM maps (\textit{p} < 0.001). Error bars were computed from 1000 bootstrapped samples of the human responses.}
    \label{fig:human_recog}
\end{figure}

We showed that the shared component of the variability across human behavioral measures is best predicted by attention maps computed using a particular class of passive attention methods, suggesting that recognition performance in both humans and machines is derived from the same visual information in images. However, this finding is based on indirect correlational evidence. In order to provide direct evidence for this claim, we tested whether human recognition performance is improved in real-time using the leading ANN maps via a speeded recognition experiment. We reasoned that recognition performance in humans will be better for images masked using their own attention maps (``correct masking'') obtained from the best ANN maps (as defined by their peak correlation to the Human PC) compared with images masked using maps obtained for different images (``incorrect masking,'' see Fig. \ref{fig:human_recog}A for the masking procedure, and Fig. \ref{fig:human_recog}B for the speeded recognition task design). Furthermore, we predicted that the difference in recognition performance on this task between these two conditions (correct masking vs. incorrect masking) would be greater if the masks are generated using the leading ANN maps than if they are generated using ANN maps that are very dissimilar to the Human PC maps. 

To test these hypotheses, we ran three additional behavioral experiments with a total of 3,600 participants recruited from AMT. Each participant viewed briefly flashed masked images (for 200 milliseconds), and had to select the best descriptors from a set of word pairs obtained from a separate labelling experiment in which a total of 160 participants took part (see Appendix). Masked images were generated by an element-wise product of a grayscale image with the map produced by the ANN model, see Fig. \ref{fig:human_recog}A. {\em Correct masking} consisted in masking an image with the map produced by the ANN for that image. {\em Incorrect masking} consisted in masking an image with the attention map produced by the ANN for different images. Representative examples of the masks along with the original color images are shown in Fig. \ref{fig:human_recog}C.

Based on the rankings of the ANN maps in terms of their peak correlations to the Human PC, we predicted that human recognition accuracy should be significantly more sensitive to the visual regions revealed by SGBP applied to the AlexNet model than those revealed by the peak maps produced by the same model using the scorecam or Grad-CAM attention methods. Paired t-tests revealed a significantly larger difference (\textit{p} < 0.001) between the $d'$ scores across map types in the correct masking condition (blue bars in Fig. \ref{fig:human_recog}D) relative to the differences in the $d'$ scores across map types in the incorrect masking condition (red bars in Fig. \ref{fig:human_recog}D; see Appendix for definition of $d'$ scores). This interaction confirms the prediction that human recognition accuracy is in fact significantly \textit{more} sensitive to the visual regions revealed by one of the models and attention methods with the highest peak correlations to the Human PC, but less sensitive to those revealed by passive attention techniques applied to a model that yielded significantly lower peak correlations to the Human PC, even though correct masking did produce a boost in recognition accuracy over incorrect masking for all three map types (\textit{p} < 0.001, Fig \ref{fig:human_recog}D). $d'$ scores broken down by individual images can be found in the SI Appendix, including the same results shown in terms of simple accuracy (\% correct) rather than $d'$ (See SI Fig. S\ref{fig:S5B}).

\vspace{-0.75mm} 

 \section{Validation experiment: human visual selectivity boosts ANN recognition}

\begin{figure}[!htb] 
    \centering
        \includegraphics[width=\linewidth]{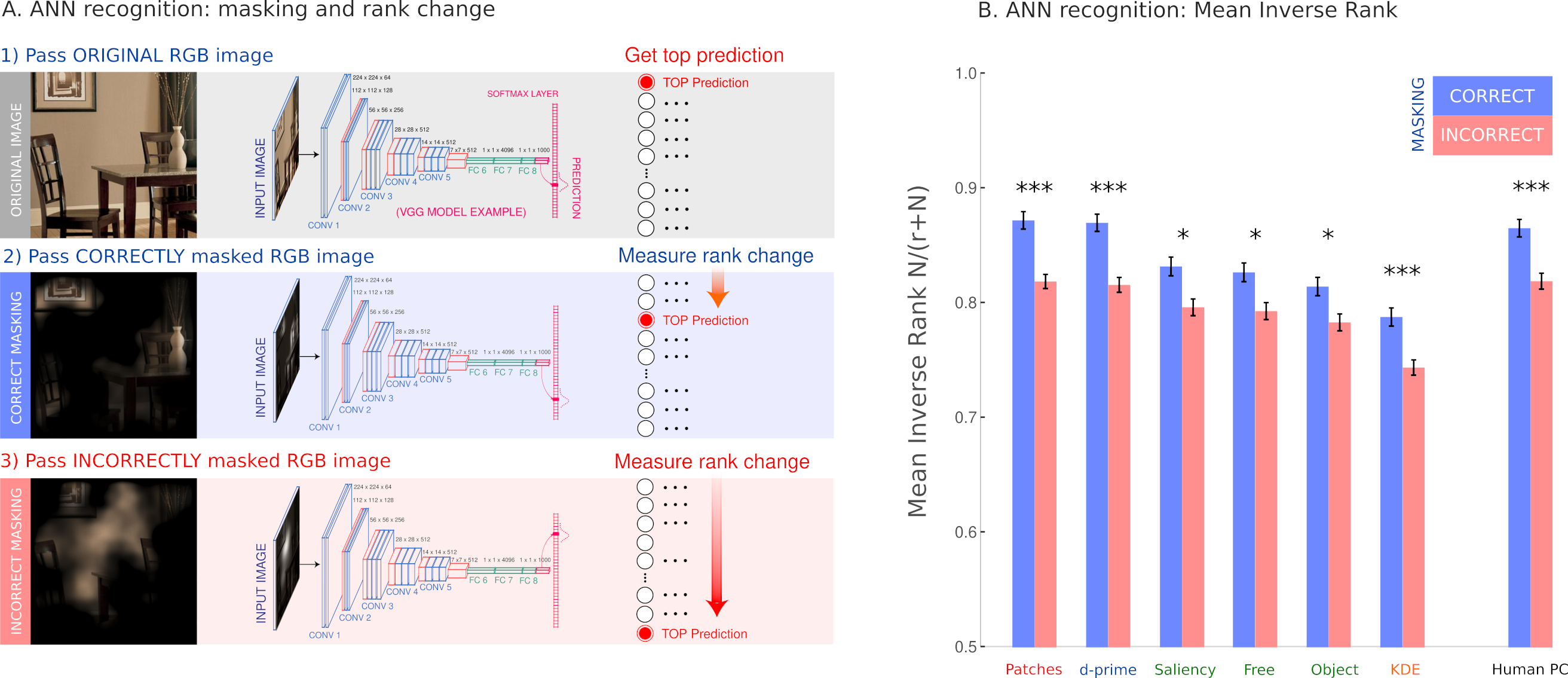} 
    \caption{ANN recognition results. A. Measuring recognition. As outlined in the text, we directly evaluate how masks derived from human maps affect ANN recognition by examining how the rank of the top image category prediction when classifying an unmasked image changes when that same image is masked with either the correct or incorrect mask. B. Mean Inverse-rank across different human maps grouped by correct vs incorrect masking, averaged across all ANN models and images. All human maps give a higher inverse-rank for correct vs incorrect masking, validating the main hypothesis that ANNs are sensitive to visual regions highlighted by human maps. We also find an interaction in the predicted direction: the effect of correct masking tended to be greater when maps were from the behavioral results that were the most highly correlated to the ANN maps overall.}
    \label{fig:ann_recog}
\end{figure}

The results of the human speeded recognition experiments suggested that we evaluate the same prediction in the opposite direction by asking whether ANN classification performance is similarly sensitive to the visual regions revealed by the human behavioral maps. To do this, we masked images using the six different human behavioral maps including the Human PC maps (Fig. \ref{fig:ANN-Human}). As with the human recognition experiments, we evaluated if ANNs show better performance in classifying correctly masked images rather than incorrectly masked images (\textit{i.e.,} images paired with the mask from a different image). Furthermore, we tested whether ANNs show improvements when the correct masks were obtained from the behavioral measures that had the highest peak correlations to the ANN maps (Fig. \ref{fig:ANN-Human}). See Fig. \ref{fig:ann_recog}A for an illustration of the whole procedure.

\textbf{Measuring recognition.} In this experiment, we evaluated how masking affects ANN recognition directly. We therefore used the ANN's classification of the original, unmasked image as a baseline, and then compared it to classification on the corresponding (correctly or incorrectly) masked image. Similar classifications in both cases would indicate good performance. We defined similarity across classifications as follows. For a given ANN, we took the top-1 category (\textit{i.e.,} the category predicted with highest confidence) for the unmasked image, and computed its rank in the predictions on the masked image ($r$). We then divided this rank by the total number of categories ($N$) to normalize for differences in the number of categories across ImageNet, CIFAR-100, and Places365 trained models. This new quantity, ($r/N$), inversely tracks recognition quality---it is lower when recognition is good, and higher when recognition is poor. To convert it into a measure that gives \textit{higher} values for better recognition and \textit{lower} values for poorer recognition (as with the human $d'$ accuracy metric), we use $N/(r + N)$ as our final measure. We refer to this measure as the \textit{inverse-rank} and computed it for all (correctly and incorrectly) masked images.

\textbf{Results.} We computed the inverse-rank across all models, for all types of human maps (Section \ref{sec:human-measures}) as well as for both kinds of masking (correct, incorrect). We found that the correctly masked images are universally more recognizable (have higher mean inverse-rank) than incorrectly masked images, across each of the different human maps (Fig. \ref{fig:ann_recog}B, $p < 0.001$). This finding validates our core prediction that ANNs should be sensitive to visual regions revealed by the human measures. 

Masked images were also more or less easy to recognize based on the human map type, and we found a significant main effect of behavioral map type on the overall mean inverse-rank (F(6,176) = 49.65, \textit{p} < 0.001) in the predicted direction: human maps that had higher peak correlations to the ANN maps also produced higher overall inverse-rank scores. We also found a significant interaction between masking condition (correct vs. incorrect) and map type, indicating that different human behavioral maps have a direct impact on recognizability (the \textit{difference} in inverse-rank between the correct and incorrect masking conditions (F(6,176) = 6.54, \textit{p} < 0.001). This finding is key, because it confirms a change in recognition performance that is predicted from the differences in overall peak correlations of the ANN maps to the different human maps. Like the interaction we observed with the human recognition experiment, it shows that behavioral measures that had higher peak correlations to the ANN maps (such as the patch rating maps) also gave higher inverse-rank score differences across masking conditions, while others, like the fixation maps, gave smaller inverse-rank score differences across masking conditions. More details of the analysis and results can be found in Appendix Fig. S6.

\section{Discussion}

Using a range of human behavioral measures, ANN models and attention techniques (Fig. \ref{fig:Human-Maps}, Fig. \ref{fig:ANN-Human}), we attempted a comprehensive examination of the similarities and differences between humans and machines with respect to their visual selectivity to image information. In a departure from prior work, we found that ANN maps are optimally predictive of a latent human visual selectivity feature that captures the joint variability between our human behavioral measures (Human PC; see Fig. \ref{fig:ANN-Human}). 

Surprisingly, simple architectures and passive guided backpropagation based attention techniques showed the highest peak correlations to the human data, and performed significantly better than the maps produced by active attention or class activation mapping based techniques. These results suggest that the same visual regions are informative to humans and machines. In addition, comparing models for which we could extract both active and passive attention maps revealed that active attention architectures consistently produce less human-like results (SI appendix section D.2). This is noteworthy because active attention models have often been conceived with the goal of emulating human attention mechanisms.

In another departure from prior work comparing human attention to ANN attention, we validated our correlational findings by running two causal experiments. In the first, we took the ANN maps and used them to mask images presented to human participants. We found that humans were better at classifying images that were masked with the correct ANN map compared with incorrect maps, and that the difference in performance between these two conditions was greater when the ANN maps used were more highly correlated to the Human PC maps (Fig. \ref{fig:human_recog}). In the second experiment, we used human maps to mask images and measured the effect on recognition performance for the ANN models. Again, we found that incorrect masking was more destructive to ANN recognition performance than correct masking (Fig. \ref{fig:Tasks}). We also found that the change in performance between the two conditions tended to be greater when the masking was done using human maps that were the most highly correlated to the ANN maps. 

Our results suggest that the regions that are discovered by attention techniques in both humans and ANNs are indeed mutually important for recognition. Moreover, these results show that we can use off-the-shelf ML methods (i.e., ANN and interpretability methods) to produce saliency maps that can predict human visual selectivity just as well as custom models (i.e. non-deep-learning visual-attention models) designed specifically for that purpose. We include predictions made by two such models: Graph-based Visual Saliency (GBVS) and Boolean Map based Saliency (BMS) \cite{harel2007graph,zhang2015exploiting} in the SI Appendix (see Fig. S\ref{fig:S2} and S\ref{fig:S4}).

One of the main findings of this paper is the fact that more human-like ANN maps tend to be more predictive of the shared component of the joint variability between human measures rather than any single human measure. Although this finding may be due in part to a reduction in measurement noise from summing the individual behavioral maps, the fact that we observed systematic variations in the visual regions that are implicated depending on the behavioral task (Fig. \ref{fig:ANN-Human}B), as well as relatively high internal reliability for each of the estimates (split-half correlations between $\rho$ = 0.63 - 0.92, see SI Appendix), suggests otherwise. 

The main limitation of this work is the inclusion of a relatively small number of images. This is due to the large number of participants needed in order to create detailed estimates for all the behavioral maps for each of the images (requiring over a hundred participants for every image). In addition, further work will be required to fully explain the factors that contribute to greater similarity between artificial attention and human visual selectivity, and ways in which we can devise systems that are better models of biological vision. In addition, while making artificial networks more human-like has practical advantages for improving their interpretability, pitfalls include introducing undesirable biases \cite{scheuerman2020we, wang2020revise, wang2020towards}). 

Overall, our results pave the way for developing new psychologically relevant benchmarks for evaluating leading ANN models, beyond comparing them to the neural basis of biological vision, or the distributions of their learned representations to the structure of human psychological representations. These results showcase new ways of combining the perspectives of machine learning and cognitive science towards developing more human-like intelligent systems.

\clearpage
\subsection*{Acknowledgements}
This work was funded by Princeton University (Grant number 1718550 from the National Science Foundation) as well as the Max Planck Society.

\setcounter{figure}{0}
\appendix
\section*{Appendix to "Passive attention in artificial neural networks \\ predicts human visual selectivity"}

\section{ANN models, passive attention, and active attention}

\begin{table}[hbt!]

In the paper we used the following naming convention.  The naming convention for passive attention maps is <architecture>-<passive attention method>-<I/C/P for ImageNet/CIFAR-100/Places365>. The naming convention for active attention maps is <attention module>-<architecture>-active-<I/C for ImageNet/CIFAR-100>.

  \caption{Human experiments}
  \label{tab:experiments}
  \centering
  \begin{tabular}{l l l l l l}
    \toprule
    Experiment                      & \makecell[l]{Map\\short name} & \makecell[l]{Approx.\\participants\\per image\\/experiment} & \makecell[l]{Total\\participants (all images\\/experiments)}  &
    \makecell[l]{New\\experiment} 
    \\
    \midrule
    Patch ratings                   & PATCH         & 9     & 225  & Yes   \\
    Discrimination accuracy         & D-PRIME       & 63    & 1,575 & Yes \\
    Spatial memory                  & SPATIAL       & 90    & 2,250 & Yes \\
    Free-viewing fixations          & FREE          & (22)    & (22) & No \cite{koehler2014saliency}\footnotemark  \\
    Saliency search fixations       & SALIENCY      & (20)    & (20)  & No \cite{koehler2014saliency}\footnotemark[\value{footnote}] \\
    Object search fixations         & OBJECT       & (19)    & (38)  & No \cite{koehler2014saliency}\footnotemark[\value{footnote}]\\
      
    Recognition validation          & N/A        & 1,200    & 3,600 & Yes  \\

    Labelling                       & N/A           & 5     & 160\footnotemark & Yes \\
    \bottomrule
  \end{tabular}
\end{table}
\addtocounter{footnote}{-1}
\footnotetext{Parentheses indicate publicly available data, see \cite{koehler2014saliency} for details. Images and eye fixation data were obtained from \url{https://data.mendeley.com/datasets/8rj98pp6km/1}.}
\stepcounter{footnote}
\footnotetext{For technical reasons, we only retained the data for 125 participants out of the full 160 that were recruited for the labelling experiment.}

\begin{table}[hbt!]
  \caption{Evaluated deep learning models}
  \label{tab:models}
  \centering
  \begin{tabular}{l l l l}
    \toprule
    Model                           & \# params     & Model ref.                    & Impl. ref. \\
    \midrule
    CIFAR AlexNet                   & 2.50M         & \cite{krizhevsky2012imagenet} & \cite{bearpaw2019pytorch} (MIT) \\
    CIFAR VGG-19 (w/ BatchNorm)     & 20.09M        & \cite{simonyan2014very}       & \cite{bearpaw2019pytorch} (MIT) \\
    CIFAR ResNet-110                & 1.73M         & \cite{he2016deep}             & \cite{bearpaw2019pytorch} (MIT) \\
    ImageNet AlexNet                & 61.10M        & \cite{krizhevsky2012imagenet} & \cite{paszke2019pytorch} (BSD-3) \\
    ImageNet VGG-16 (w/ BatchNorm)  & 138.37M       & \cite{simonyan2014very}       & \cite{paszke2019pytorch} (BSD-3) \\
    ImageNet ResNet-101             & 44.55M        & \cite{he2016deep}             & \cite{paszke2019pytorch} (BSD-3) \\
    ImageNet EfficientNet-B0        & 5.29M         & \cite{tan2019efficientnet}    & \cite{wightman2019timm} (Apache 2.0) \\
    ImageNet ViT-S/16               & 22.05M        & \cite{dosovitskiy2020image, steiner2021train} & \cite{wightman2019timm} (Apache 2.0) \\
    Places AlexNet                  & 58.50M        & \cite{krizhevsky2012imagenet} & \cite{zhou2017places} (CC BY) \\
    Places ResNet-50                & 24.26M        & \cite{he2016deep}             & \cite{zhou2017places} (CC BY) \\
    \midrule
    CIFAR LTPA VGG                  & 19.99M        & \cite{jetley2018learn,simonyan2014very}   & \cite{saoyan2020learn} (GPLv3) \\
    CIFAR ABN ResNet-110            & 3.06M         & \cite{fukui2019attention,he2016deep}      & \cite{fukui2020attention} (MIT) \\
    CIFAR ABN DenseNet-BC ($L=100$, $k=12$) & 1.12M & \cite{fukui2019attention,huang2017densely} & \cite{fukui2020attention} (MIT) \\
    ImageNet ABN ResNet-101         & 62.58M        & \cite{fukui2019attention,he2016deep}      & \cite{fukui2020attention} (MIT) \\
    \bottomrule
  \end{tabular}
\end{table}

\begin{enumerate}[wide, labelindent=0pt]
  \item \textbf{ANN models.} We evaluated several different CNN models, some with active attention modules and some without. Some of these models were trained on CIFAR-100 \cite{krizhevsky2009learning}, meaning that they take in 32x32 images as input and classify into 100 possible classes; some were trained on ImageNet 2012 \cite{deng2009imagenet}, meaning that they take in 224x224 images as input and classify into 1000 possible classes; and some were trained on Places365-Standard \cite{zhou2017places}, meaning that they take in 224x224 images as input and classify into 365 possible classes. The particular models we evaluated are listed in Table~\ref{tab:models}. The models in the top part of the table were evaluated with passive attention methods, and the models in the bottom part of the table have active attention modules.
  \item \textbf{Passive attention methods.} Passive attention methods have been developed to provide insights into which parts of an input image a model is attending to. In addition to early gradient-based techniques \cite{simonyan2013deep,springenberg2014striving}, we used class activation mapping techniques \cite{selvaraju2017grad,wang2020score}. We adapted open-source PyTorch implementations of these methods \cite{uozbulak2019pytorch}\footnote{Note that we made some modifications to this code, and these modifications are present in our supplementary materials. In particular, we fixed the SmoothGrad implementation so that $\sigma$ is computed correctly, and we generalized the guided backpropagation implementation (and therefore all the other methods that rely on guided backpropagation) so that it works for architectures other than AlexNet and VGGNet. Our code is available at \url{https://github.com/czhao39/neurips-attention}.} (released under the MIT License). For each method, we computed attention maps with respect to the top class prediction made by the model. For Grad-CAM, Score-CAM, and CAMERAS, activation maps were derived from the last convolutional layer. We describe each of the methods below:
  \begin{enumerate}
     \item Guided backpropagation (GBP) \cite{springenberg2014striving} -- This computes an imputed version of the gradient. It is the same as standard backpropagation except it prevents the backward flow of negative gradients by zeroing them out. By doing so, it uses higher layers to ``guide'' backpropagation to lower layers.
     \item Guided gradient $\times$ image (GBPxIM) \cite{shrikumar2016not} -- This is the same as the output of guided backpropagation except it is multiplied by the (normalized) input image. Roughly, this gives a first-order approximation of the effect of setting any given input pixel to 0.
     \item SmoothGrad with guided backpropagation (SGBP) \cite{smilkov2017smoothgrad} -- This method attempts to address the noisiness in raw gradient visualizations. The authors posit that this noisiness is because the gradient may fluctuate sharply at small scales, which seems plausible especially given that, due to ReLU activation functions, the output generally is not even continuously differentiable. To address this, SmoothGrad adds Gaussian noise to the original input and performs guided backpropagation, repeats this to generate a sample of sensitivity maps, and then averages these together to produce the final sensitivity map. We used a sample size of 30, and we set $\sigma$ such that the noise level is 10\%.
    \end{enumerate}
    The next three methods are based on class activation maps (CAMs) \cite{zhou2016learning}. The original CAM method relies on a global average pooling (GAP) layer between the final convolutional layer and the fully-connected layer in an image classification CNN. A GAP layer simply computes the average value over each activation map in the final convolutional layer, and uses these averages as inputs into the fully-connected layer. The CAM is then defined as the linear combination of these final activation maps, weighted by the target class's weights in the fully-connected layer. This CAM indicates the discriminative regions of the image used by the CNN to identify that class. Since not all CNN architectures have a GAP layer, several closely related techniques have been developed that do not rely on a GAP layer. We describe the three methods we used below:
    \begin{enumerate}[resume]
     \item Grad-CAM \cite{selvaraju2017grad} -- This stands for gradient-weighted class activation mapping. This is the same as standard CAM except that instead of relying on a GAP layer, Grad-CAM uses gradients to weigh each activation map.
     \item Score-CAM \cite{wang2020score} -- There are several issues with gradient-based approaches like Grad-CAM, such as gradients being noisy and tending to vanish. Therefore, Score-CAM avoids gradients altogether. Score-CAM is the same as standard CAM except that instead of relying on a GAP layer, Score-CAM uses forward-passing scores to weigh each activation map. An activation map's forward-passing score for a given class is defined as the model's score for that class when the input image is masked by the activation map.
     \item CAMERAS \cite{jalwana2021cameras} -- CAM methods generally rely on the low-resolution activation maps of the final convolutional layer, resulting in saliency maps that may be imprecise. CAMERAS addresses this by up-/down-sampling the input image to multiple resolutions (we used resolutions of 100x100, 224x224, and 1000x1000) and running each of these through the ANN. The resulting gradients and activations are averaged and then used to produce the final, fused class activation map.
  \end{enumerate}
  We used each of the above passive attention methods to acquire attention maps from each of the models in the top part of Table~\ref{tab:models}.
 \item \textbf{Active attention methods.} Relatively recently in the area of computer vision, some models have been developed which have active attention modules built into their architectures, so that these models explicitly attend to certain locations in the input. We evaluated two different active attention modules, which we describe below:
  \begin{enumerate}
      \item Learn to Pay Attention (LTPA) \cite{jetley2018learn} -- We refer to this architecture by the title of its paper, LTPA. In particular, we used the (VGG-att3)-concat-pc model defined in the paper, and apply attention before the max-pooling layers. The model is based off of a VGGNet architecture, with three attention estimators at intermediate layers within the CNN. Of the three attention estimators, we used the attention maps produced by the middle estimator in our analyses.
      \item Attention Branch Network (ABN) \cite{fukui2019attention} -- ABN also uses an end-to-end trainable attention module. ABN consists of three modules: feature extractor, attention branch, and perception branch. The feature extractor and perception branch are constructed by splitting a baseline CNN into two parts. The attention branch is placed after the feature extractor and is based on class activation mapping. We used the resulting CAM as the attention map in our analyses.
  \end{enumerate}
  We evaluated the LTPA and ABN models listed in the bottom part of Table~\ref{tab:models}.
 \end{enumerate}
 
\section{Non-ANN models of human attention}
The focus of this work is to evaluate visual selectivity of ANNs in relation to human visual selectivity. In particular, we do \textit{not} claim that ANN saliency represents the state-of-the-art in modeling humans. Nevertheless, it is still instructive to compare ANN saliency to non-ANN models of human attention (i.e., models designed specifically to predict human attention). We examined two state-of-the-art non-ANN models of human attention:
\begin{enumerate}
    \item Graph-Based Visual Saliency (GBVS) \cite{harel2007graph} -- GBVS first computes standard biologically-inspired feature maps (Gabor filters, contrast maps, and luminance maps), then uses a graph-based approach to compute activation maps on these feature maps which are meant to highlight locally ``unusual" regions, then uses a graph-based approach to normalize these activation maps in a way that concentrates the mass on these maps, and finally sums these together. This approach is naturally parallelizable, suggesting biological plausibility.
    \item Boolean Map based Saliency (BMS) \cite{zhang2015exploiting} -- As opposed to most models which focus on properties of local image patches (e.g., contrast and rarity), BMS tries to model a global perceptual phenomenon found to be relevant to human visual attention---figure-ground segregation. In particular, BMS computes a saliency map by detecting surrounded regions in an image.
\end{enumerate}

\section{Human visual selectivity estimation}

\subsection{Experimental design and procedure}

\textbf{``Informativeness'' patch ratings task.} 
We used the procedure described by \cite{henderson2018meaning,henderson2017meaning} to generate dense ``meaning'' maps for all 25 images (taken from the database of images used by \cite{koehler2014saliency} for which detailed eye-movement fixation patterns were available). For the patches, we extracted circular image regions from a 12 by 12 regular grid over the entire image. The patches were extracted from high-resolution versions of the images that were full-color 2430 by 2430 pixel images. The diameter of the patches was 442 pixels (see SI Appendix Fig. \ref{fig:S1}A). During the experiment, we presented each of the patches along with a small thumbnail of the full image that included a green circular marker over the image to indicate where the patch was extracted from, for context. Participants rated the ``informativeness or recognizability'' of the image content revealed by each of the patches using a Likert scale (1 = ``Very low recognizability'', 2 = ``Low recognizability'', 3 = ``Somewhat low recognizability'', 4 = ``Somewhat high recognizability'', 5 = ``High recognizability'', 6 = ``Very high recognizability''), see SI Appendix Fig. \ref{fig:S1}A). In the experiment, the terms ``informativeness'' and ``recognizability'' were used interchangeably. Participants rated all 144 patches for a given image per experiment, and we obtained judgments from 9 unique participants for each image patch over AMT. Participants were paid \$$2$ for their participation. The exact instructions at the start of the experiment were as follows: ``In the task, you will see a circular image patch along with a thumbnail of the full image from which the patch was taken to provide context. A circle over the full thumbnail image will indicate the location of the patch. your job is simply to rate the content revealed inside each circular image patch (NOT the full image) in terms of how RECOGNIZABLE or INFORMATIVE it is using a 6-point Likert scale (‘very low’, ‘low’, ‘somewhat low’, ‘somewhat high’, ‘high’, ‘very high’). There is no right or wrong answer.'' SI Appendix Fig. \ref{fig:S1}D shows the final map results for this task, for all 25 images.

\textbf{Change sensitivity discrimination task.} We used the same design and procedure described in \cite{langlois2021serial}. We started by producing a regular grid of possible point locations that spanned the full area of each of the images (see SI Appendix Fig. \ref{fig:S1}B). The grid points were 7 pixels apart. During the task, participants saw an image presented for 1000 ms with a red point displayed over it (SI Appendix Fig. \ref{fig:S1}B). Following a 1000 ms blank delay, the image reappeared with the point either in the same exact location relative to the image or in a shifted position. In the ``shifted'' condition, the point was shifted by 6 pixels somewhere along a circular radius around the original point location, sampled at random. The second display remained for 1000 ms on the screen and was followed by a 2AFC (``red dot same'', or ``red dot shifted''). Participants could take as long as they liked to choose a response, although they had to complete the experiment within one hour before the experiment expired on AMT. We obtained responses from a total of 9 participants for each grid point, and for each condition (``same'' or ``shifted''). The full instructions at the start of the experiment were as follows: ``In this experiment, you will see two images presented one after the other. These images will have a red dot placed over them. Your task is to determine if the red dot is in the same spot relative to the image for both images in the pair, or if the red dot appears displaced the second time it is presented. NOTE: The displays will be shown in random positions on the screen, even in cases when the red dot is placed in the EXACT SAME spot over the image! So part of the challenge is to ignore the random shifting of the overall display, and focus on the RELATIVE positions of the dots in relation to the images, ignoring the random overall displacements.'' For the discrimination experiment, compensation was \$1.5, and included 120 trials. Participants could take part in as many discrimination experiments as they wished.

\textbf{Spatial memory task.} We used the same design and procedure introduced by \cite{langlois2021serial} to estimate spatial memory priors. Participants were presented with an image with a red point initialized somewhere over the image for 1000 ms. The location of the point was sampled from a uniform distribution. Participants were instructed to reproduce the exact location of the point relative to the image from memory as accurately as possible. Exact instructions were: ``In this task, a background image display with a red dot over it will be shown for some time, and will be followed by a blank screen. The background image will reappear but without the red dot inside. Your task will be to place the dot in the exact position where you previously saw it (relative to the background)!''. Overall positions of the displays, including the point and image, were shifted by a random horizontal and vertical offset between 0 and 80 pixels on the screen canvas (in order to avoid a strategy of marking the absolute positions of the points on the screen). The response of the participant was sent to another participant on AMT who performed the same task, with their response becoming the stimulus for the next participant (and so on). A total of twenty iterations (generations) of the process were completed for each chain (There was a total of 250 chains for each image, beginning as random point location initializations over the image). We ended each experiment after approximately 12 hours. Typical participation included 105 trials, and the average time needed to complete the task was about 12-14 minutes.  A typical experiment included about 90 participants. Compensation for taking part in the task was dependent on performance (between \$1.4 and \$1.5). Participants could take part only once per experiment, but could take part in more than one experiment (for more than one image). We only retained the chains that reached the full twenty iterations. Participants completed 10 practice trials prior to moving on to the 95 experimental trials.

\subsection{Cross-participant reliability in the behavioral measures}
We evaluated the internal reliability of our behavioral estimates using split-half reliability. We computed 100 random splits of the data for each human behavioral task and measured the average split-half correlations for the 100 split-half pairs with the Spearman-Brown correction. The results were: $\rho$ =.85 for the patch ratings estimates, $\rho$ =.88 for free fixations, $\rho$ =.92 for cued object search fixations, $\rho$ =.87 for saliency search fixations, $\rho$ =.63 for the spatial memory KDEs, and $\rho$ =.75 for the $d^{\prime}$ discrimination accuracy maps.

\subsection{Map estimation}

\begin{figure}[!htb] 
    \centering
        \includegraphics[width=\linewidth]{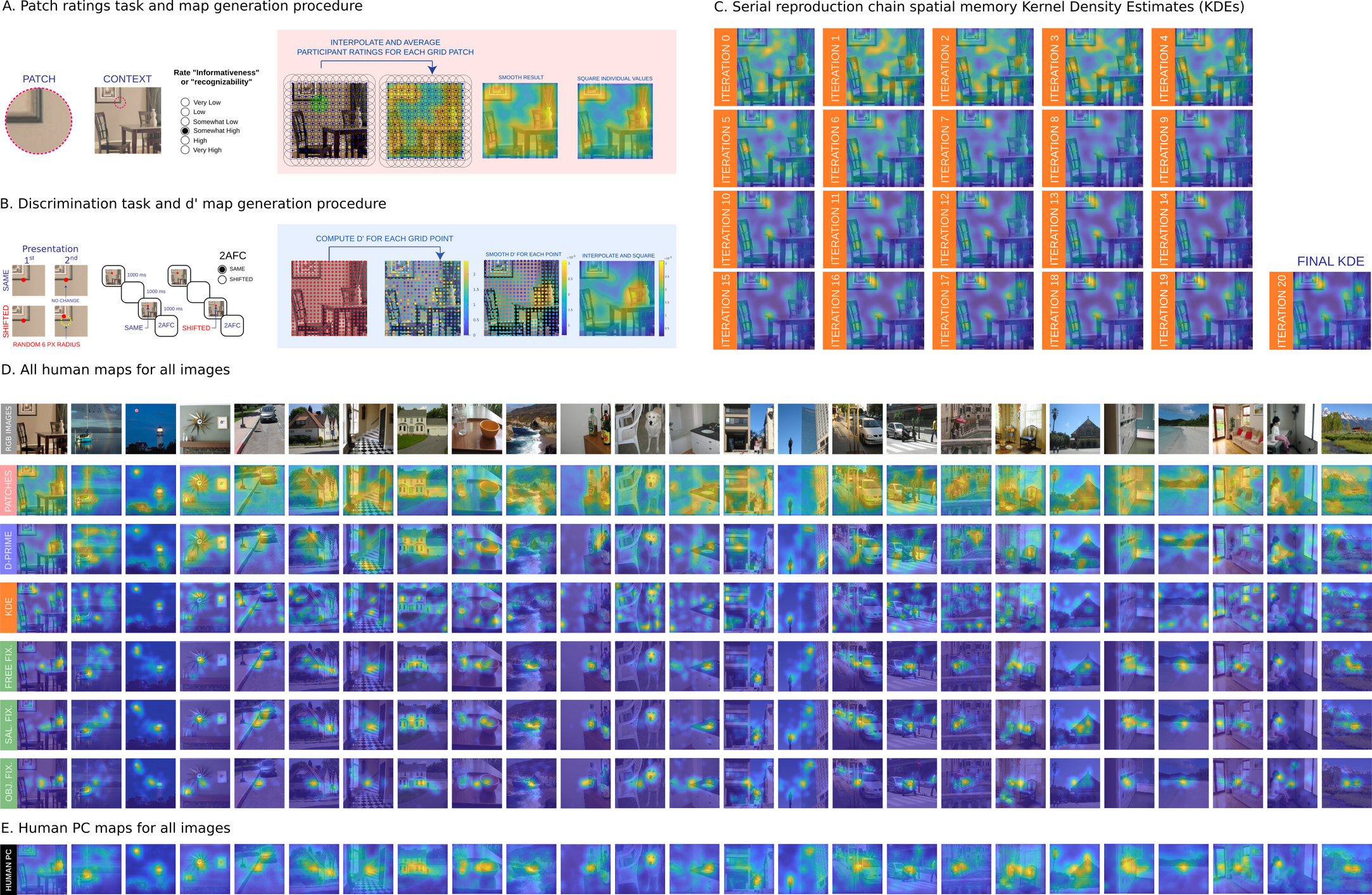}
    \caption{Human behavioral tasks, map generation procedures, and results for all images. A. ``Informativeness'' patch ratings task, and map estimation. For each pixel, maps were obtained by averaging the ratings given for patches that overlapped with that pixel, including ratings given for neighboring locations weighted by an isotropic Gaussian centered at the pixel. B. Discrimination task and map generation procedure. Participants completed 2AFC tasks for each grid point sampled from a regular grid of locations over the image. For each point, we computed $d'$ using the 2AFC responses, smoothed the grid $d'$ values, and interpolated between them to obtain continuous change sensitivity estimates over the entire images. For both experiments we exponentiated the final map values by squaring each value to increase the dynamic range. C. Spatial memory localization estimates and serial reproduction procedure. Participants were instructed to remember precise point locations from memory. Each response from a given iteration in the chain was forwarded to the next participant in the subsequent iteration. Kernel Density Estimates (KDEs) of the results at each of the chain iterations are shown, including the final (20th) estimate. D. Behavioral map results for all tasks, including the 3 fixations (see \cite{koehler2014saliency} for details). E. Human PC maps for all images.}
    \label{fig:S1}
\end{figure}

\textbf{``Informativeness'' patch map estimation.} To compute full maps from the patch ratings, we averaged the ratings made for all patches at a given pixel location and ratings made at neighboring pixel locations weighted by an isotropic Gaussian kernel (as a weighted interpolation of the ratings at a given pixel with the ratings at neighboring pixel locations). To increase the dynamic range, we then exponentiated the resulting map by squaring each of the individual values in the map to obtain the final maps. SI Appendix Fig. \ref{fig:S1}D shows the final results for all the images.

\textbf{Change sensitivity map estimation.} We obtained $d^{\prime}$ values for each of the discrimination grid points by using the 2AFC responses obtained for each grid point (see formula for computing $d'$ included in Appendix \ref{sec:d-prime}). We then convolved the grid of raw $d^{\prime}$ values with a fixed Gaussian kernel. Next, we generated full continuous $d^{\prime}$ map estimates by interpolating between the grid points using cubic interpolation. Finally, to increase the dynamic range, we exponentiated the final result by squaring each of the map values. SI Appendix Fig. \ref{fig:S1}B shows the results including the raw $d^{\prime}$ grid point values, the smoothed $d^{\prime}$ grid point values before the interpolation, and the smoothed $d^{\prime}$ interpolated maps (final discrimination accuracy maps) for one of the images. SI Appendix Fig. \ref{fig:S1}D shows the final interpolated change sensitivity estimates for all the images.

\textbf{Spatial memory Kernel Density Estimation (KDE).} We computed KDEs using the data from the last (20th) iteration of the serial reproduction chains. For each point in the 20th iteration, we computed a Gaussian kernel centered at that point with a diagonal covariance matrix. We then summed all of the Gaussian kernels and then normalized to get the final KDE. Final KDEs are shown in SI Appendix Fig. \ref{fig:S1}D.

\textbf{Free fixations, cued object search fixations, and saliency search fixations.} To compute maps of the fixations data, we used the raw eye-movements data in \cite{koehler2014saliency}, which consisted in (x,y) coordinates of the fixations over the images, and used the same kernel density estimation procedure used for the spatial memory task (see above) to obtain the final maps. These are shown in SI Appendix Fig. \ref{fig:S1}D.

\begin{figure}
    \centering
        \includegraphics[width=\linewidth]{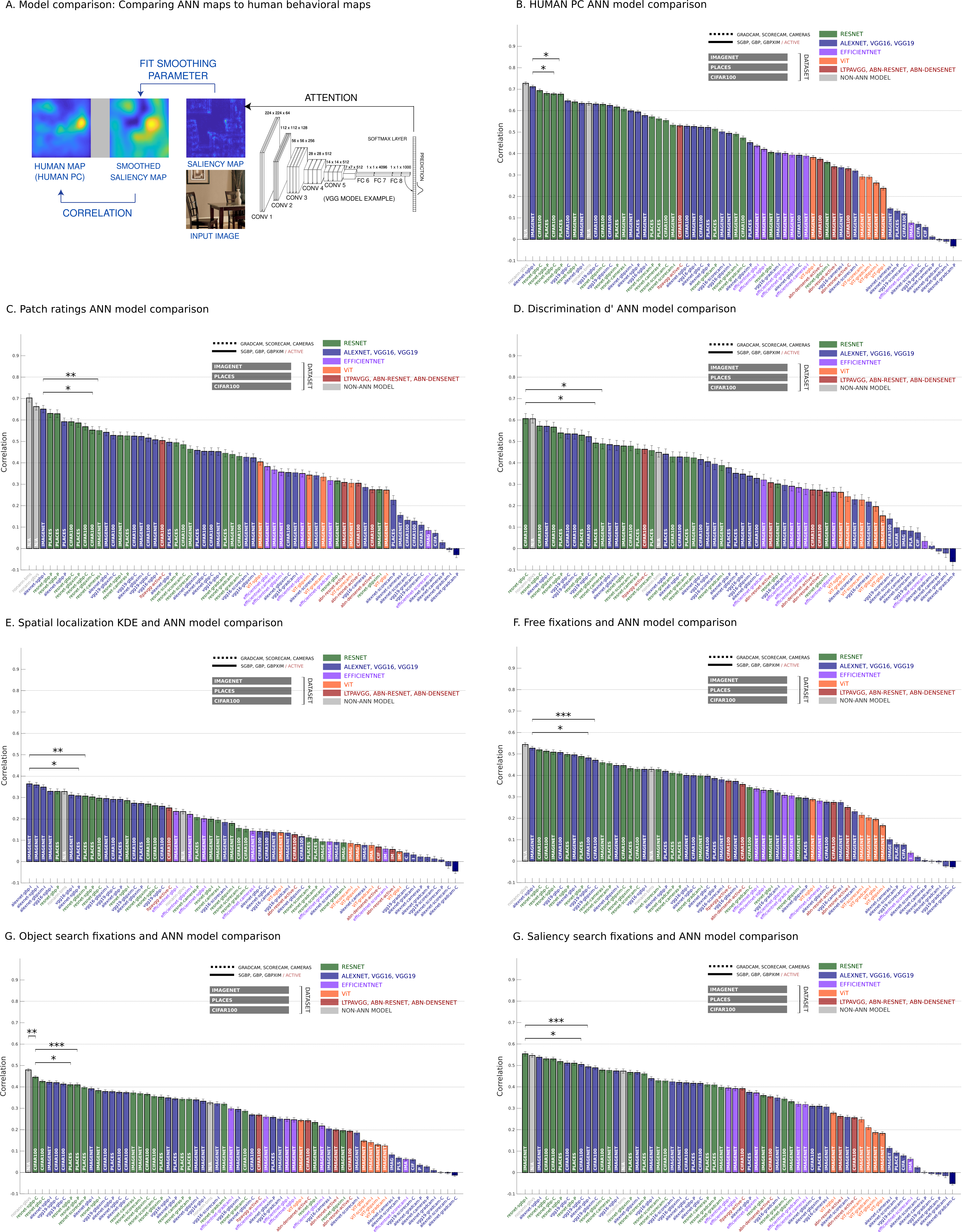}
    \caption{ANN attention and human visual selectivity. A. Schematic of the ANN and human map comparison procedure. We fit a smoothing parameter that maximized the average correlation (across all 25 images) of the smoothed ANN maps to the corresponding human maps. The schematic illustrates an example for a single image, correlating a smoothed ANN attention map with the corresponding human PC map. B. Sorted peak average correlations across all 25 images for all ANN maps to the human PC maps. The maps produced by ImageNet pretrained AlexNet probed using Smooth Guided Backpropagation (SGBP) were significantly more highly correlated to the human maps than most of the remaining ANN maps (\textit{p} < 0.001; with the Bonferroni correction for multiple comparisons applied). C. results for the ``informativeness'' patch ratings maps. As with the Human PC results, the same ANN maps outperformed the remaining ANN maps (\textit{p} < 0.001). Panels D-H show the results for all remaining human maps. They show similar results, with guided backpropagation methods producing ANN maps that were more predictive than others overall (\textit{p} < 0.001).}
    \label{fig:S3}
\end{figure}

\section{Comparing human and ANN maps}

\subsection{ Principal Component Analysis (PCA) for estimating Human PC}
In order to estimate the shared component of the variance between all the human behavioral maps, we completed a Principal Component Analysis (PCA) with the concatenated human maps as input (a 250,000 x 6 matrix containing 6 vectorized versions of the concatenations of the maps across all 25 100 x 100 images, for each map type). We z-scored (standardized) each of the 6 2,500 x 100 human map concatenations prior to vectorizing them, and then concatenated them into the full 250,000 x 6 matrix for the input to the PCA. We used \textsc{MATLAB's} \textit{pca} function to obtain the 6 PCA coefficients (loadings) for each of the human map vectors that explain the maximal shared variance between them. We then obtained the Human PC as a linear combination of the 6 human map types using these factor loadings. Because we did not want to overrepresent the  three human fixations maps (which were highly intercorrelated (\textit{r} = 0.72-0.86) we downweighted each by multiplying them by a factor of $\sqrt{\frac{1}{3}}$ to obtain the final PC map. The PC map images for all 25 images are shown in SI Appendix Fig. \ref{fig:S1}E. The first PC explains 53.3\% of the variance in the human behavioral measures (whereas the next three components explain 19.9\%, 12.4\% and 6.8\% of the variance).

\begin{figure}[!htb] 
    \centering
        \includegraphics[width=\linewidth]{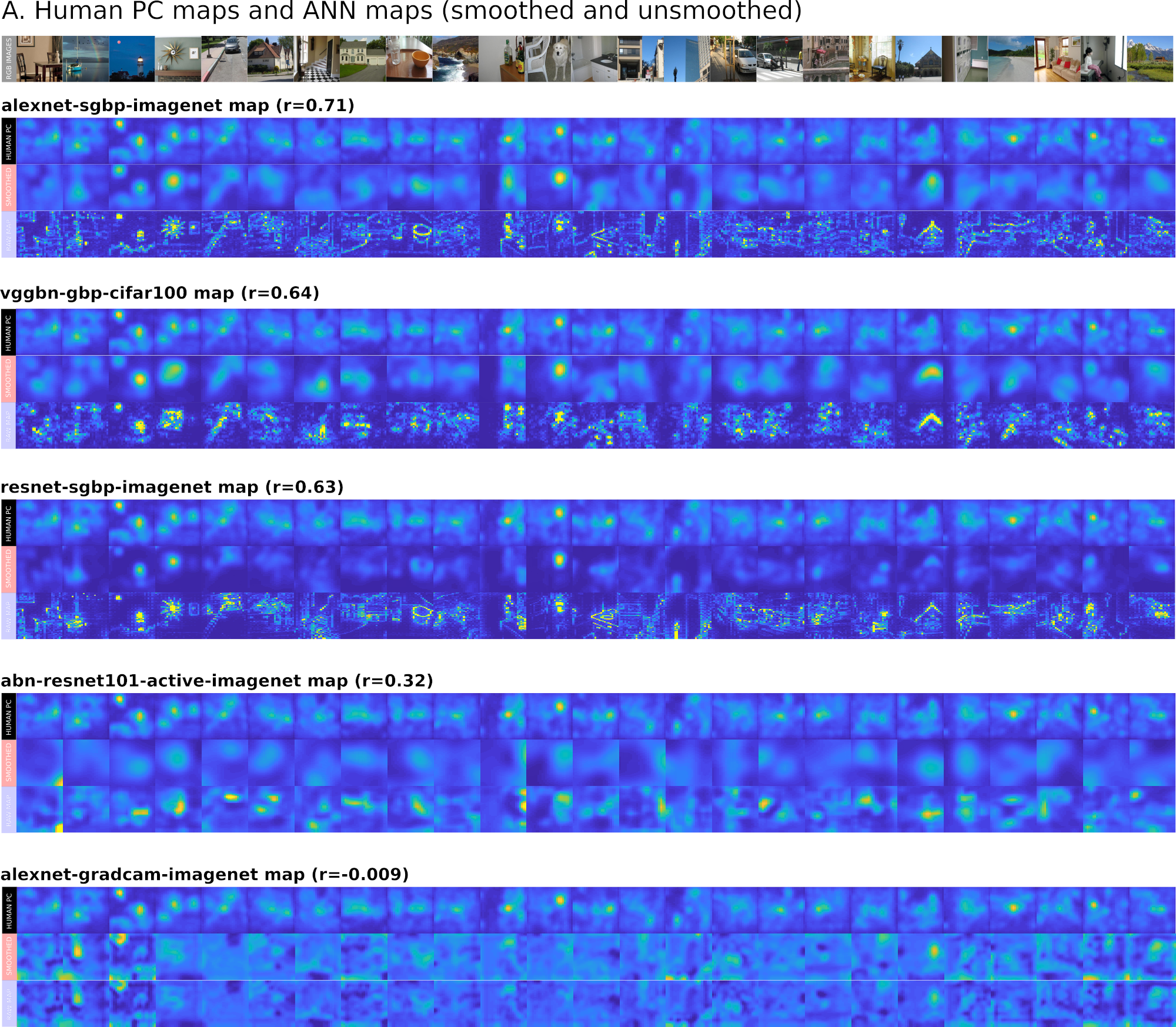}
    \caption{Human PC maps, and ANN maps (optimally smoothed and raw unsmoothed examples). A. Human PC maps (first row of each group of three). Second row in each group of three shows optimally smoothed ANN maps, and third row in each group shows the unsmoothed raw ANN maps, for all images. Examples for some of the best performing ANN maps are shown, including intermediate and worse performing examples. The peak average \textit{r} correlations to the human PC maps achieved with optimal smoothing are shown for each ANN map type.}
    \label{fig:S2}
\end{figure}

\begin{figure}[!htb] 
    \centering
        \includegraphics[width=\linewidth]{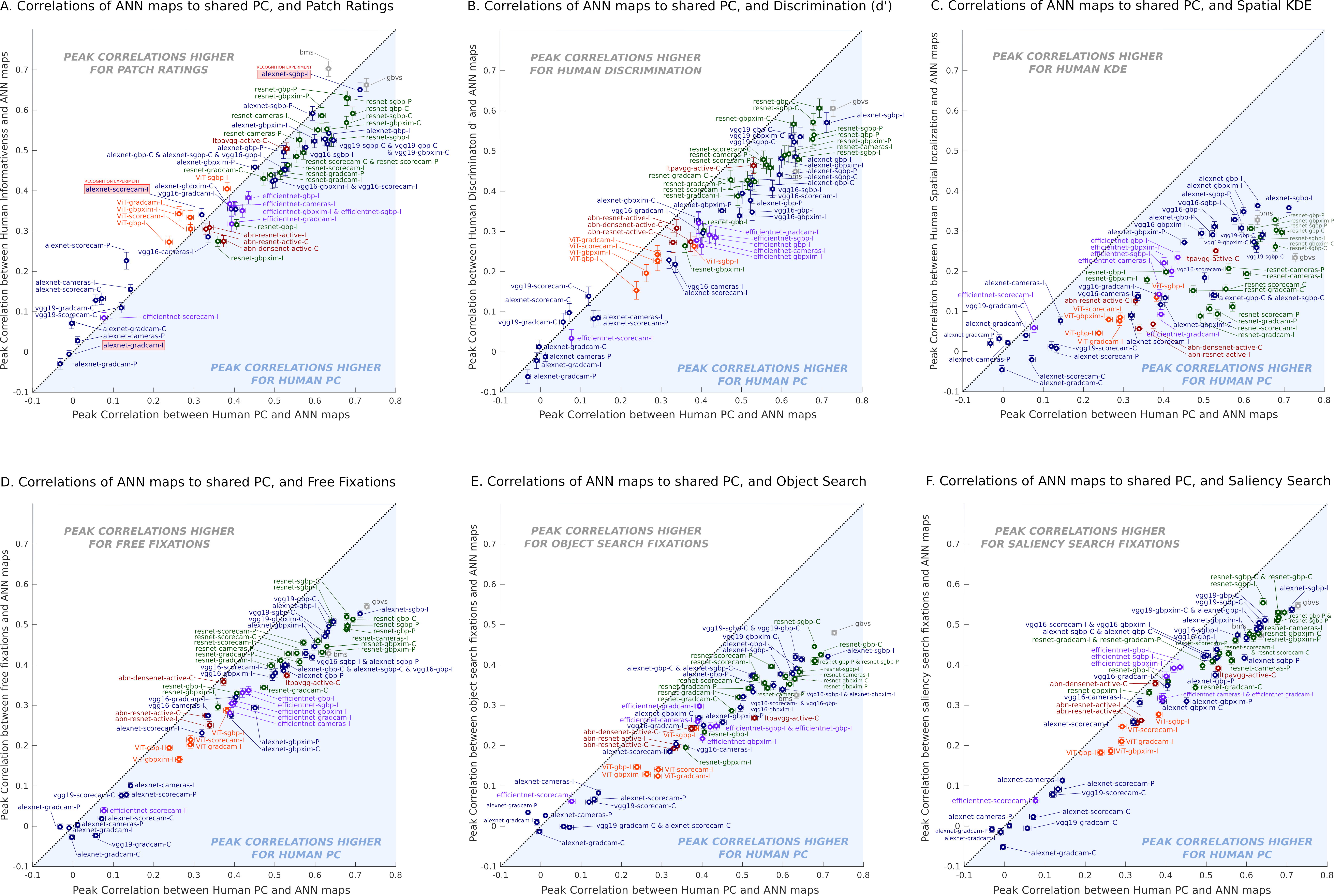}
    \caption{Peak correlations between all ANN maps and Human PC (x-axis) and peak correlation between all ANN maps and each of the human behavioral maps (y-axis). Error bars were estimated from 100 bootstrapped samples of the human data. A. Results for the patch ratings experiment shown in Fig. \Fc B in the main text. Points plotted below the dotted diagonal line inside the blue shaded area correspond to ANN map types that achieved peak correlations to the Human PC maps that were higher than the corresponding peak correlations of the same ANN maps to the particular human map types shown. Results for discrimination accuracy $d'$ maps, spatial memory KDE maps, and the three eye-movement fixation maps are shown in panels B-F. The fact that peak average correlations achieved by the higher-performing ANN maps to the Human PC maps are overall higher than those achieved by the same ANN maps optimized to each individual human map indicates that the most human-like ANN maps capture a shared component of the variability across all human behavioral measures rather than the unique variance captured by any one of the individual behavioral maps. Note, we also include peak correlations achieved by non-deep learning computational attention methods in each subplot. These are the peak correlations for Graph-based Visual Saliency (GBVS) and Boolean Map based Saliency (BMS) methods. These results show that we can use off-the-shelf ANN and interpretability methods to produce saliency maps that can predict human visual selectivity just as well as custom models (i.e. non-deep-learning visual-attention models) designed specifically for that purpose.}
    \label{fig:S4}
\end{figure}

\subsection{Predicting human PC and behavioral maps using ANN attention}

For each human map type (including the human PC), we computed the peak average correlation between the 25 image maps (which were resized to 100 x 100 images) and the corresponding attention maps produced by each of the ANN methods. For each ANN method, we searched over a smoothing parameter range of $\sigma = 0-30$ and selected the one that produced the peak average correlation (over all 25 images) of the ANN method maps to the corresponding human maps by smoothing each using \textsc{MATLAB's} \textit{imgaussfilt} function (see SI Appendix Fig. \ref{fig:S3}A). Barplots showing the ANN methods sorted by peak average correlation to each of the human behavioral maps are shown in SI Appendix Fig. \ref{fig:S3}B-H. Error bars for each of the ANN methods shown in each barplot were obtained by resampling the human data with replacement and estimating 100 new human maps, and then taking the standard deviation of the averaged correlations of each of the 100 bootstrapped map estimates to the optimally smoothed ANN method maps across all 25 images. For the human PC, the 100 bootstrapped map estimates were obtained by a linear combination of the 100 bootstrapped sample map estimates obtained for each of the 6 behavioral maps using the weights (loadings) from the PCA (see section above for PCA details). Pairwise t-tests comparing the peak ANN method maps to all other ANN method maps revealed significantly higher correlations for the maps produced by AlexNet pretrained on ImageNet using SGBP relative to most of the remaining ANN methods for human maps except the object search fixation maps (\textit{p} < 0.001; We applied the Bonferroni correction for multiple corrections; see SI Appendix Fig. \ref{fig:S3}). Qualitative examples of the ANN maps, including unsmoothed and smoothed examples (fit to the human PC maps), are shown in SI Appendix Fig. \ref{fig:S2}A. Examples shown include ANN maps that were among the top predictors of the human PC maps, as well as the lowest predictors of the human PC maps (SI Appendix Fig. \ref{fig:S2}A).

Overall average peak correlations achieved between the smoothed ANN maps and the human PC maps were higher than the peak correlations achieved between the ANN maps optimized to each individual human map. This finding is illustrated in Fig. \Fc B of the main text for the human patch ratings results, and for all 6 human behavioral maps in SI Appendix Fig. \ref{fig:S4}. This indicates that the ANN maps capture a shared component of the variability across all human behavioral measures rather than the unique variance captured by any one of the individual human maps.

With respect to passive and active attention, we observed more human-like results using passive attention techniques applied to the same models for which we could extract active attention maps. In particular, we found that LTPA (which is a VGGNet model with an added active attention module, and is trained on CIFAR-100) achieved a peak correlation with the human PC of \textit{r} = 0.53, while SGBP applied to VGG19 (also trained on CIFAR-100) achieved a higher peak correlation of \textit{r} = 0.64 (\textit{p} < 0.001). Similarly, ABN ResNet-110 (which is a ResNet-110 model with an added active attention module trained on CIFAR-100) achieved a peak correlation of \textit{r} = 0.31, while SGBP applied to ``vanilla'' ResNet-110 produced a much higher peak correlation of \textit{r} = 0.68 (\textit{p} < 0.001). Finally, ABN ResNet-101 (trained on ImageNet) achieved a peak correlation of \textit{r} = 0.32, while SGBP applied to ResNet-101 (also trained on ImageNet) has a higher peak correlation of \textit{r} = 0.63 (\textit{p} < 0.001). These results indicate that comparing identical discriminator networks for which we could obtain both passive and active attention maps shows that active attention networks produce far less human-like results.

\subsection{Statistical comparison of averaged correlations}
When comparing averaged correlations between different methods or training sets, we computed the averaged peak correlations for each condition and for each image. For each comparison, we then performed a paired t-test across all 25 images for the two averaged vectors. When multiple comparison were reported, we used the Bonferroni correction.

\section{Human recognition experiments: design}

\begin{figure}[!htb] 
    \centering
        \includegraphics[width=\linewidth]{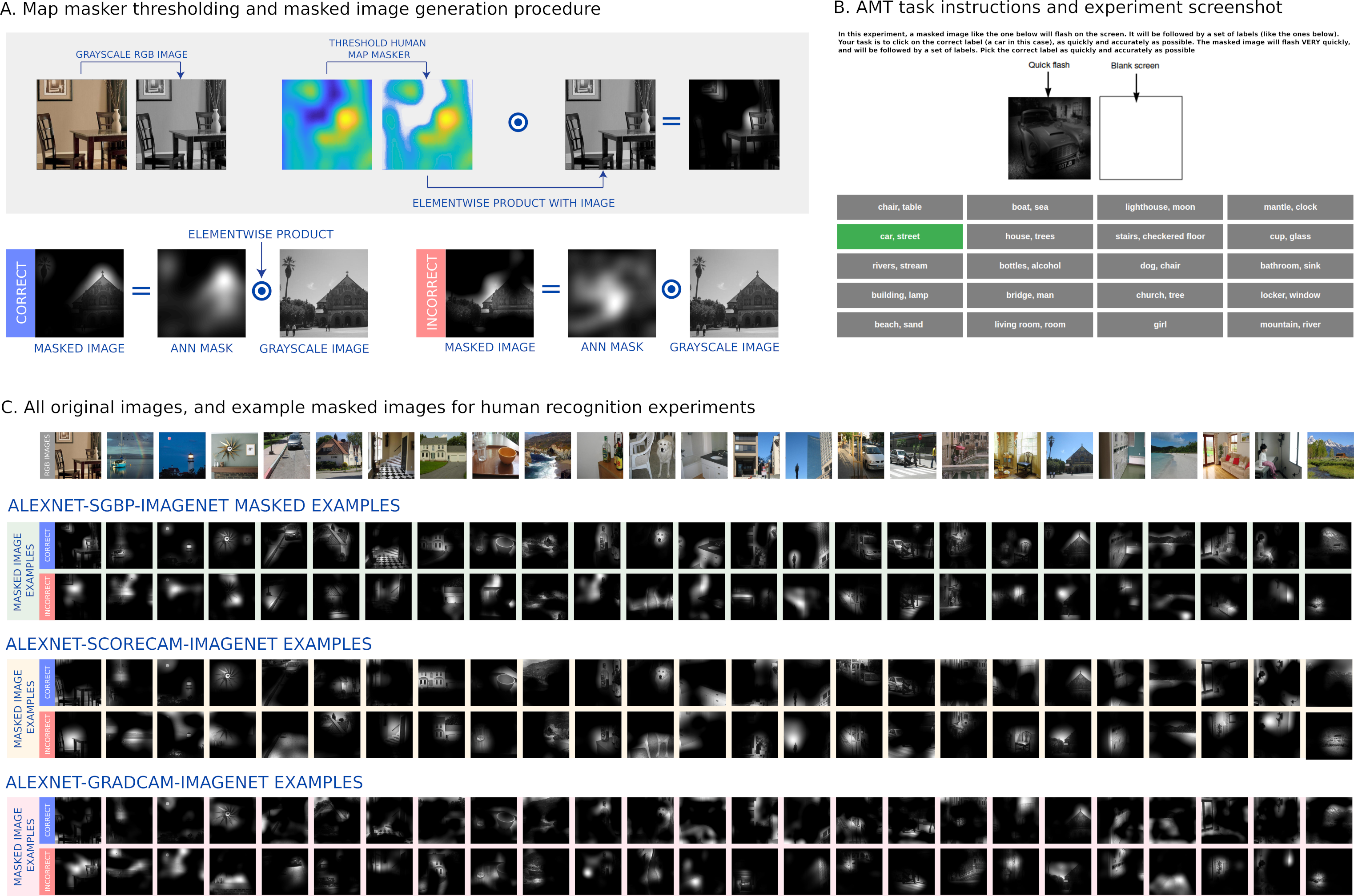} 
    \caption{Human recognition experimental design and procedure. A. Masking procedure. Grayscale versions of the original images were masked by an elementwise product with smoothed ANN maps that were thresholded in order to reveal 50\% of the total image area. Correct masking consisted in using the smoothed ANN map obtained for the same image as a masker, while incorrect masking consisted in using an ANN map obtained from a different image, with a random rotation applied. Only incorrect maps that were relatively weakly correlated to the correct mask (less than \textit{r} = 0.4) were retained for incorrect masking. B. Screenshot of the Amazon Mechanical Turk (AMT) experiment instructions. C. All the original RGB images, and representative examples of correctly masked and incorrectly masked images for each experiment.}
    \label{fig:S5A}
\end{figure}

\begin{figure}[!htb] 
    \centering
        \includegraphics[width=\linewidth]{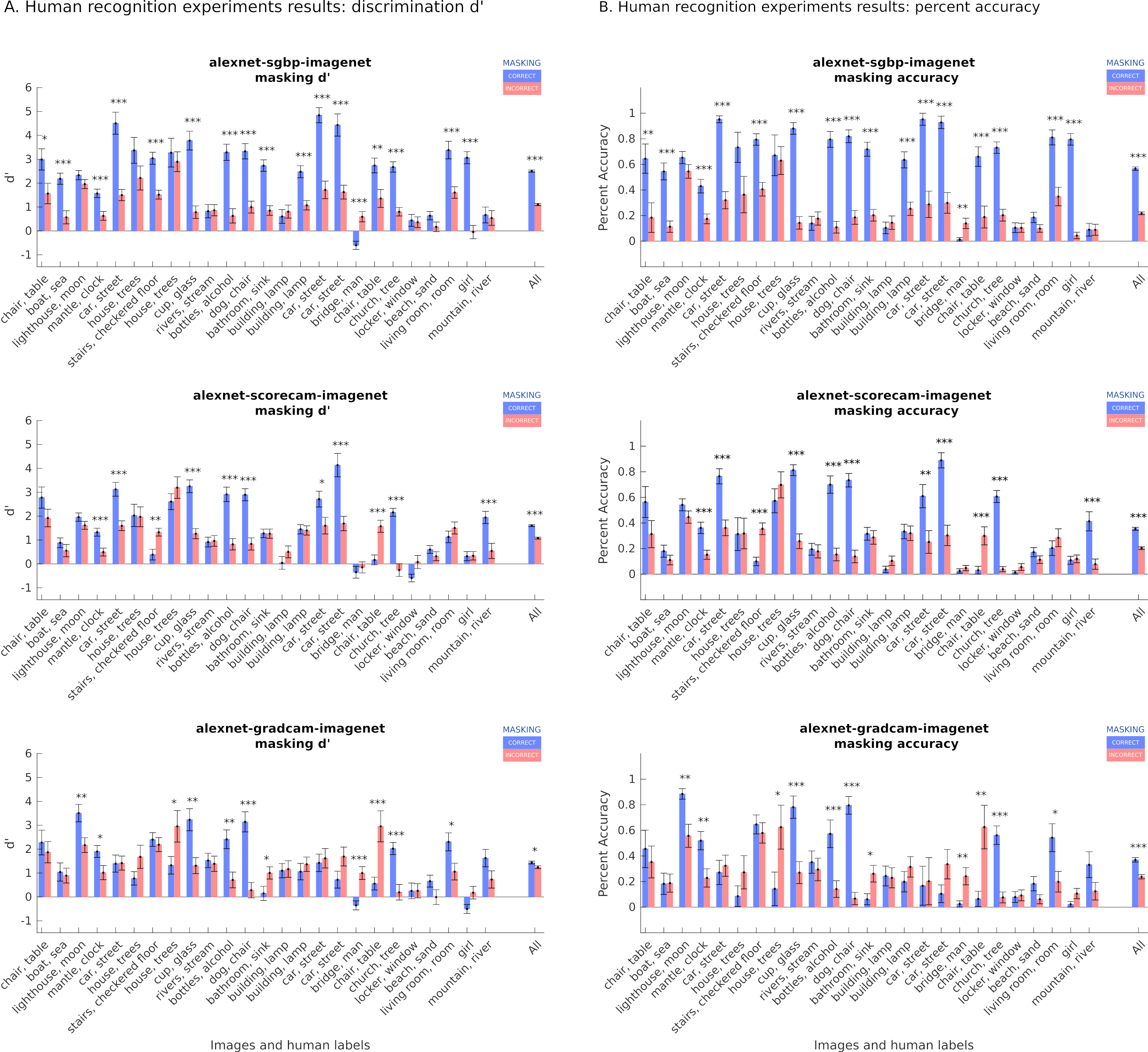} 
    \caption{Human recognition experiment results. A. $d'$ results for masked images obtained using the SGBP method on AlexNet pretrained on ImageNet (first row). Also shown are $d'$ results for masked images obtained using the Score-CAM and Grad-CAM methods on AlexNet pretrained on ImageNet (second and third rows, respectively). Results show a clear main effect of correct vs. incorrect masking for most individual images (\textit{p} < 0.001; with the Bonferroni correction for multiple comparisons applied). A significant interaction (\textit{p} < 0.001) confirms that correctly masked images obtained using SGBP yielded higher $d'$ than correctly masked images obtained using the Score-CAM or Grad-CAM methods (\textit{p} < 0.001; with the Bonferroni correction for multiple comparisons applied). Results are shown for participants that had a higher than 15\% accuracy during the task. B. The same results are shown in terms of percent accuracy instead of $d'$.}
    \label{fig:S5B}
\end{figure}

\subsection{Human recognition experiment: experimental design}

Participants completed a speeded categorization task for 10 images (trials). The 10 trials were comprised of a unique permutation of a multiset of 5 correctly masked image cases, and 5 incorrectly masked image cases. The 10 images were a set of unique images sampled without replacement from the full set of 25 images. Each set did not contain more than one image containing the same object label (labels were obtained through crowdsourcing, see below). For example, a set could only contain one car and street image. We generated 200 sets of 10 images with 5 correctly masked cases and 5 incorrectly masked cases each, ensuring that the total number of correctly masked and incorrectly masked cases was equalized for each image. This produced a total of 2000 unique masked images (See SI Appendix Fig. \ref{fig:S5A}C for representative examples, for each of the 25 images).

The masking was done as follows. The image was converted to grayscale (so as to avoid participants using color cues). The mask was an ANN map smoothed using the smoothing parameter that yielded the peak correlation to the human data. We started by thresholding the map to ensure that only 50\% of its values were above 0 (as a way of roughly equalizing the amount of image area being revealed under the mask across images and conditions). We then multiplied the thresholded map and grayscale image elementwise to obtain the final masked image (see SI Appendix Fig. \ref{fig:S5A}A and C).

Each participant on Amazon Mechanical Turk (AMT) completed 10 experimental trials. Their instructions were as follows: ``In the task, you will see an image flash for about 200 milliseconds. Your job is simply to select the best caption (set of object words) from a list of words that will be presented to you. Choose the list of words that provide the best description of the image you saw. The images will be partly masked, and may be difficult to see. If you are unsure, just provide your best guess.'' See SI Appendix Fig. \ref{fig:S5A}B for a screenshot of the experiment instructions from the AMT task. Participants completed 4 practice trials prior to completing the full 10 experimental trials from the image set assigned to them.

\subsection{Crowdsourcing image captions for the human recognition experiment}
We recruited a total of 160 participants but retained usable responses for 125. For each image, 5 participants were instructed to do the following ``Please write three words or short phrases that summarize the contents of the image. If someone were to see these three words or phrases, they should understand the subject and context of the image.'' Participants were paid \$0.2 to label all 25 images. To come up with the final list of labels for the human recognition experiments, we tallied the frequency of all the labels provided by all participants, and retained the top-one or top-two most frequent word labels as final labels.  
 
\section{Human recognition experiments: analysis}

\subsection{Calculating \texorpdfstring{$d^{\prime}$}{Lg}}
\label{sec:d-prime}
$d^{\prime}$ scores were computed for each image, and for each condition (\textit{correct} vs. \textit{incorrect} masking) by calculating the False Alarm ($\mathrm{FA}$) rate (the number of times a given label set was selected when the image shown was not an instance of that label set, over the number of times that the presented images were not instances of that label set), and the $\mathrm{HIT}$ rate (the number of times that a given label set was selected when the image shown was an instance of that label set, over the number of times that all the presented images were instances of that label set). $d^{\prime}$ is given by: $ d'= Z(\mathrm{HIT}) - Z(\mathrm{FA})$ where the function $ Z(p), p \in [0,1] $, is the inverse of the cumulative distribution function of the Gaussian distribution.

\subsection{Testing the effect of masking and ANN map type}

For each image and for each ANN map type (see SI Appendix Fig. \ref{fig:S5A}C), we computed $d'$ using the formula described in the previous section. For each of the 25 images we computed $d'$ using $\mathrm{HIT}$ and $\mathrm{FA}$ rates from participant choices on the nAFC task. SI Appendix Fig. \ref{fig:S5B}A and B show the results broken down by image, for all recognition experiments. Fig. \ref{fig:S5B}A shows the results in terms of $d'$. While some images were intrinsically harder to recognize than others (with lower $d'$ scores in the correct masking conditions), the results reveal clear effects of correct vs. incorrect masking for most individual images (\textit{p} < 0.001; we applied the Bonferroni correction for multiple comparisons). Fig. \ref{fig:S5B}B shows the same results in terms of percent accuracy. In order to mitigate the problem of negative $d'$ values, we excluded data from participants that had a lower than 15\% accuracy in their trials.

Paired t-tests revealed a significant difference (\textit{p} < 0.001) between the $d'$ scores across models for the correct masking condition (blue bars in Fig. \Fd D of the main text and in the far right of each barplot in SI Appendix Fig.  \ref{fig:S5B}A). This finding confirms the prediction that human recognition accuracy is in fact significantly \textit{more} sensitive to the visual regions revealed by one of the models with the highest peak correlations to the human PC maps, but less sensitive to those revealed by passive attention techniques on a model with a much lower peak correlation to the human PC maps, even though correct masking did produce a boost in recognition accuracy over incorrect masking for both map types (\textit{p} < 0.001, Fig. \Fd D). 

\begin{figure}[!htb] 
    \centering
        \includegraphics[width=\linewidth]{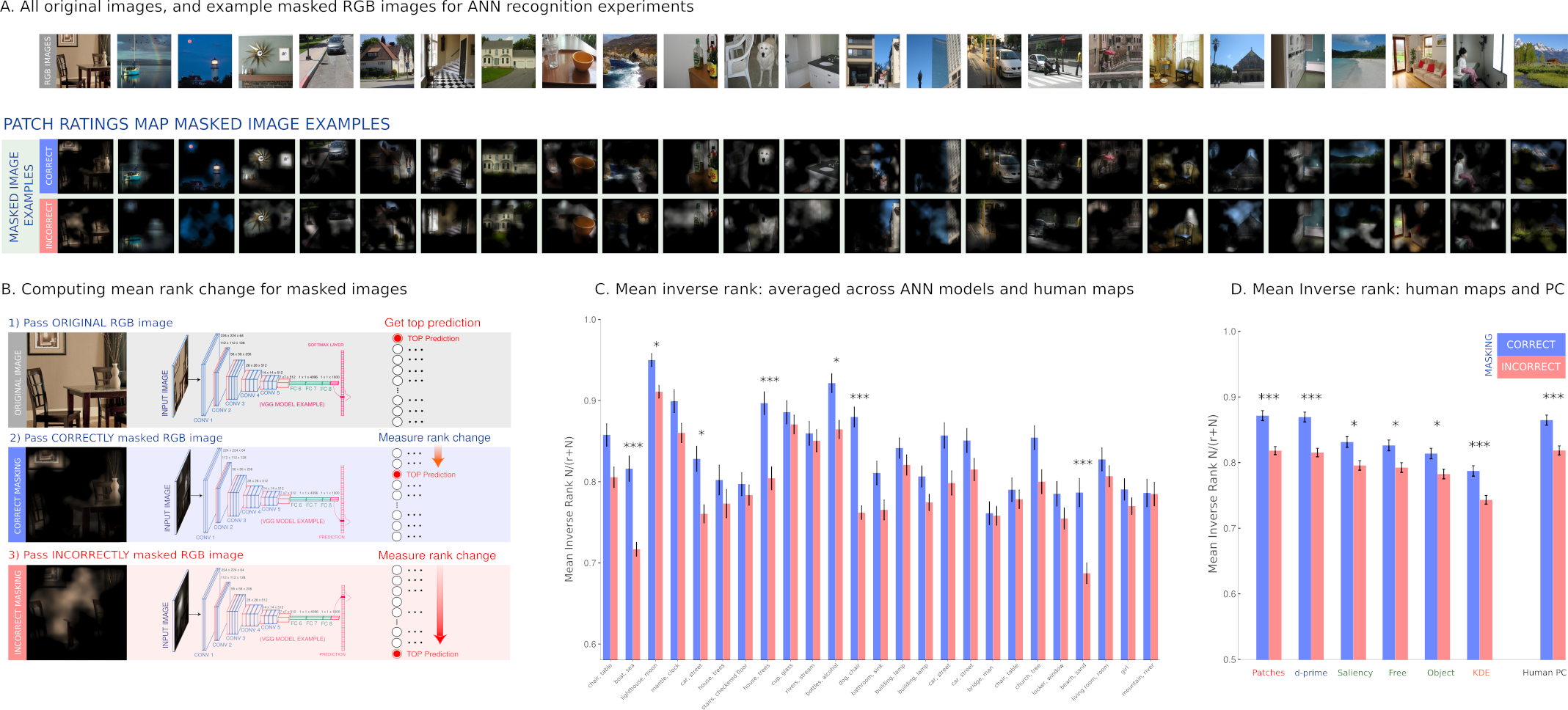}
    \caption{ANN recognition. A. All original RGB images, and masked RGB image examples. Masked image examples shown were obtained using the human patch ratings maps. B. Procedure for computing the change in rank of the top prediction. For each model, the top-1 prediction on the original unmasked image was taken as the ground truth label. We measured the rank distance (change in rank) of this prediction for the same image when correctly masked and incorrectly masked. We inverted the rank distance for comparability to the $d'$ metric used for the human experiments. We call this inverted quantity the \emph{inverse-rank}. C. Mean inverse-rank results averaged across all ANN models and human behavioral maps. Results show clear trends for most images and significant effects (\textit{p} < 0.05) for the lighthouse image, as well as the boat, house, dog, and beach images \textit{p} < 0.001. We applied the Bonferroni correction for multiple comparisons. To help with comparing the ANN and human recognition results, the x-axis labels in the barplot are the labels provided by human participants, and not predictions made by the ANN models. They are ordered in the same order as the full RGB images in A. D. Mean inverse-rank result for each human behavioral map, including the Human PC factor.}
    \label{fig:S6}
\end{figure}

\section{ANN recognition experiments: design}

\subsection{Masking procedure} We produced masked images using the same procedure used for the human recognition experiments as outlined in the main text (in the section called "Validation Experiment: ANN Visual Selectivity Boosts Human Recognition"), except that we used RGB (instead of grayscale) images in order to minimize distribution shift for the ANNs (See SI Appendix Fig. \ref{fig:S6}A for examples of masked RGB images). We expected a reduction in classification accuracy on masked images because of the significant distribution shift caused by masking the RGB images, regardless of the mask type. As a consequence, we simply examined the \textit{difference} in accuracy when an image is masked with a correct mask versus with an incorrect mask. For each image, we produced one correctly masked image, and then generated 24 incorrectly masked images by pairing the image with masks of the same map type (such as using the patch ratings maps) from the other 24 images; this gave us a total of $25 \times 25$ masked images for each of the 25 images in the dataset.

\subsection{Inverse rank computation}
For each model, the top-1 prediction on the unmasked image is taken as the ground truth label. The rank distance (change in rank) of this prediction due to (correct or incorrect) masking captures how a mask affects ANN recognition on the given image; for example, if the top-1 predicted label on the unmasked image attains a rank position of 5 on a masked image, the rank distance is 4. SI Appendix Fig. \ref{fig:S6}B shows a schematic of the rank distance measurement procedure. As described in the main text (Section "Validation Experiment: Human Visual Selectivity Boosts ANN Recognition" in the paper), this rank distance is inverted to align with trends in the $d'$ metric described in Appendix \ref{sec:d-prime}, and we call the inverted quantity the \emph{inverse-rank}. We computed inverse-rank as $N/(r + N)$, where $N$ is the total number of categories, and $r$ is the rank of the top-1 category predicted by the model for the full (unmasked) RGB image.  

\section{ANN recognition experiments: analysis}

We computed the inverse-rank across all models, for all types of human maps (Section "Validation Experiment: ANN Visual Selectivity Boosts Human Recognition"), and across all 25 different masks for each of the 25 images (see SI Appendix Fig. \ref{fig:S6}C for results broken down by image). We found that the correctly masked images are more recognizable (have higher inverse-rank) than incorrectly masked images for all types of human behavioral maps (error bars represent $95\%$ confidence intervals across the 25 images). This supports the hypothesis that ANNs are sensitive to visual regions highlighted by human behavioral maps. Fig. \ref{fig:S6}D shows the results for each human behavioral map, including the Human PC map.

We performed a subsequent ANOVA analysis using the masking condition (2: correct vs. incorrect), the specific image (25), and the behavioral map type (7) as repeated measures. We find a significant effect of masking condition (F(1,176)=184.85, \textit{p} < 0.001) indicating that the correct masking condition gives higher mean inverse-rank. This trend is seen even across individual images (averaged across the 7 behavioral maps) in Fig.~\ref{fig:S6}. We also find a significant effect of the type of behavioral map (F(6, 176)=49.65, \textit{p} < 0.001); \textit{i.e.,} certain types of behavioral maps produce more recognizable masks overall. For example, images masked using Human PC, patch ratings, and discrimination accuracy maps were more recognizable overall than those masked using spatial localization maps (KDEs), across both correct and incorrect masking conditions. This might be driven by the statistics of the masks and their compatibility with ANN classification. We also note that these differences are consistent with the correlational findings: Overall, the average of the peak correlations across all ANN maps were higher for the Human PC, patch ratings, and discrimination accuracy maps, than they were for the spatial localization (KDE) maps.

Finally, we examine whether different behavioral maps also capture different information, and whether this affects ANN recognizability. This is reflected in whether different behavioral maps vary in how well they distinguish between correct and incorrect masking for ANN recognition. We indeed find a significant interaction between masking condition (correct vs. incorrect) and the kind of behavioral map (F(6,176)=6.54, \textit{p} < 0.001). This indicates that the difference in inverse-rank between correct and incorrect masking varies significantly across different behavioral maps. We find a trend that maps from behavioral measures that have higher overall peak correlation to ANN attention maps (Human PC, patch ratings, and discrimination accuracy maps) also lead to higher differences in inverse-rank across masking conditions, while measures such as eye movement patterns (that have lower overall peak correlation with ANN attention maps) give lower mean inverse-rank differences. 

\begin{figure}[!htb] 
    \centering
        \includegraphics[width=\linewidth]{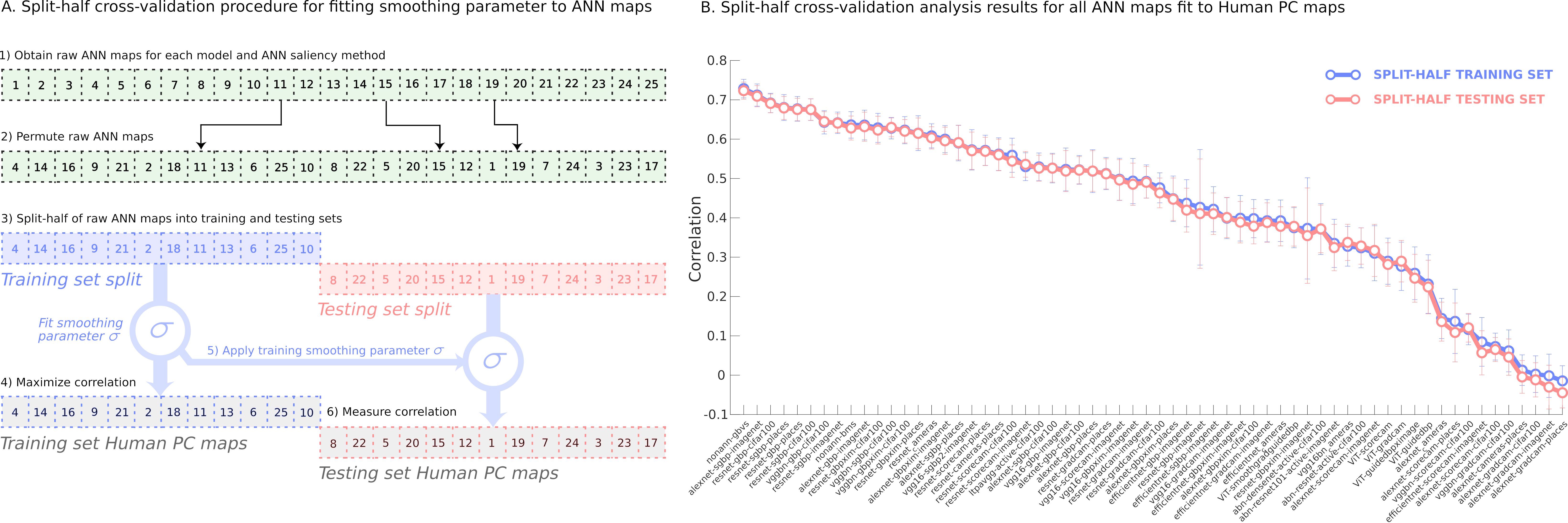}
    \caption{Split-half cross-validation analysis. A. Procedure. For each raw ANN map, we randomly permuted all the maps and divided them into a training and separate testing set. Next, we optimized a smoothing parameter $\sigma$ that maximized the average correlation of the maps in the training set to the corresponding human PC maps. We then measured the average correlation achieved between the ANN naps in the testing set and their corresponding human PC maps when smoothed using the $\sigma$ parameter fit to the training set maps. We repeated this process 100 times for each ANN map type. B. Results. The results are essentially unchanged relative to the results reported in the main text, where no cross-validation was done, and the smoothing parameter was fit to all the data. Error bars reflect 100 random permutations and splits of the ANN maps for each ANN map type. Results are ordered in terms of the peak average correlations achieved for the training set.}
    \label{fig:S7}
\end{figure}

\section{Split-half cross-validation analysis of smoothing parameter}

As a supporting analysis, we estimated peak correlations between the ANN maps and the human PC maps using split-half cross-validation when estimating the smoothing parameter (a Gaussian smoothing kernel with standard deviation $\sigma$). SI Appendix Fig. \ref{fig:S7}A illustrates the procedure. For each ANN map, we repeated the following process 100 times: We started by randomly permuting the order of the maps and splitting them into training and testing sets (the training set contained 12 maps, and the testing set contained the remaining 13 maps). Next, we fit the smoothing parameter $\sigma$ to the training set by maximizing the average correlation between the training set ANN maps and the corresponding human PC maps. We then evaluated the correlation achieved between the 13 testing set ANN maps and the corresponding human PC maps when smoothed using the same $\sigma$ parameter.

Results of the analysis are shown in SI Appendix Fig. \ref{fig:S7}B. They show a near total overlap between the peak correlations achieved for the training set maps, and those obtained for the testing set maps. In fact, the results are essentially unchanged relative to those reported in the main text: performance of smoothed test set maps using smoothing parameters $\sigma$ fit to the training set maps produced nearly identical ranges in peak correlations to the human PC (between \textit{r} = 0.73 and \textit{r} = -0.01 for the training set, and between \textit{r} = 0.72 and \textit{r} = -0.04 for the testing set). In addition, we observe a nearly identical rank order in the peak correlations to the human PC (for instance, across 100 random splits of the data, we observed an average correlation of \textit{r} =.711 (sd =.028) for the peak correlation of the AlexNet SGBP maps to the human PC maps in the training set, and an average of \textit{r} =.710 (sd =.026) for the peak correlation of the AlexNet SGBP maps to the human PC maps in the testing set when applying the smoothing parameter optimized to the training set). Note that we have also included non-ANN attention-based model maps in the analysis (GBVS \cite{harel2007graph} and BMS \cite{zhang2015exploiting}), for reference.

\bibliography{main}

\end{document}